\documentclass[10pt,twocolumn,letterpaper]{article}

\usepackage{iccv}
\usepackage{times}
\usepackage{epsfig}
\usepackage{graphicx}
\usepackage{cite}
\usepackage{array}
\usepackage[american]{babel}

\makeatletter		

\newcommand{\Rmnum}[1]{\expandafter\@slowromancap\romannumeral #1@}
\makeatother

\usepackage{amssymb,amsmath,amsthm}
\usepackage{booktabs}
\usepackage{multirow}
\usepackage{makecell}
\usepackage{bm}		
\usepackage{xcolor}
\usepackage[colorlinks, citecolor=blue]{hyperref} 
\usepackage{threeparttable}
\usepackage{subfigure}

\usepackage{algorithm}
\usepackage{algorithmicx}
\usepackage{algpseudocode}
\usepackage{tabularx}
\usepackage{multirow}
\usepackage{dsfont}

\usepackage[title]{appendix}

\newcolumntype{L}[1]{>{\raggedright\arraybackslash}p{#1}}
\newcolumntype{C}[1]{>{\centering\arraybackslash}p{#1}}
\newcolumntype{R}[1]{>{\raggedleft\arraybackslash}p{#1}}

\renewcommand{\algorithmicrequire}{\textbf{Input:}}
\renewcommand{\algorithmicensure}{\textbf{Output:}}

\def\onedot{.}
\def\eg{\emph{e.g}\onedot}
\def\ie{\emph{i.e}\onedot}
\def\etal{\emph{et al}\onedot}

\iccvfinalcopy 

\def\iccvPaperID{2308} 

\ificcvfinal\fi

\begin{document}

\title{Fast Full-frame Video Stabilization with Iterative Optimization}

\author{Weiyue Zhao$^{1}$\hspace{0.2in}
        Xin Li$^{2}$\hspace{0.2in}
        Zhan Peng$^{1}$\hspace{0.2in}
        Xianrui Luo$^{1}$\hspace{0.2in}
        Xinyi Ye$^{1}$\hspace{0.2in}
        Hao Lu$^{1}$\hspace{0.2in}
        Zhiguo Cao$^{1}$\footnotemark[1]\hspace{0.1in}
        \\
$^1${Key Laboratory of Image Processing and Intelligent Control, Ministry of Education; School of Artificial} \\
{ Intelligence and Automation, Huazhong University of Science and Technology, Wuhan 430074, China} \\
$^2$ Department of Computer Science, University of Albany, Albany NY 12222 \\ 
{\tt\small \{zhaoweiyue, peng$\_$zhan, xianruiluo, xinyiye, hlu, zgcao\}@hust.edu.cn, \,}
{\tt\small xli48@albany.edu}
\vspace{-2mm}
}

\maketitle
\ificcvfinal\thispagestyle{empty}\fi

\renewcommand{\thefootnote}{\fnsymbol{footnote}} 
\footnotetext[1]{Corresponding author}

\begin{abstract}
Video stabilization refers to the problem of transforming a shaky video into a visually pleasing one. The question of how to strike a good trade-off between visual quality and computational speed has remained one of the open challenges in video stabilization. Inspired by the analogy between wobbly frames and jigsaw puzzles, we propose an iterative optimization-based learning approach using synthetic datasets for video stabilization, which consists of two interacting submodules: motion trajectory smoothing and full-frame outpainting. First, we develop a two-level (coarse-to-fine) stabilizing algorithm based on the probabilistic flow field. The confidence map associated with the estimated optical flow is exploited to guide the search for shared regions through backpropagation. Second, we take a divide-and-conquer approach and propose a novel multiframe fusion strategy to render full-frame stabilized views. An important new insight brought about by our iterative optimization approach is that the target video can be interpreted as the fixed point of nonlinear mapping for video stabilization. We formulate video stabilization as a problem of minimizing the amount of jerkiness in motion trajectories, which guarantees convergence with the help of fixed-point theory. Extensive experimental results are reported to demonstrate the superiority of the proposed approach in terms of computational speed and visual quality. The code will be available on \href{https://github.com/zwyking/Fast-Stab}{GitHub}.
\end{abstract}

\begin{figure*}
  \centering
  \includegraphics[width=0.98\textwidth]{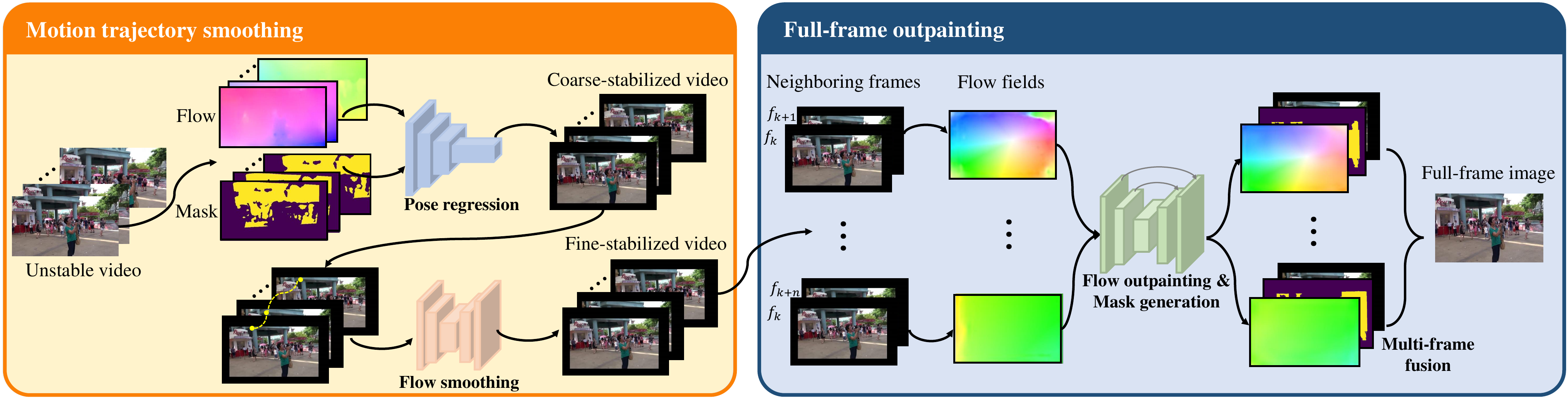}
  \caption{\textbf{Overview of our video stabilization framework.} It consists of motion trajectory smoothing (in Sec.~\ref{sec:stab_optim} and Sec.~\ref{sec:video stab}) and full-frame outpainting modules (in Sec.~\ref{sec:video outpainting}). The former adopts the two-level (coarse-to-fine) stabilizing algorithm to obtain a stabilized video. The latter further render a full-frame video with strategies of flow outpainting and multiframe fusion.} \vspace{-0.2in}
  \label{fig:fig_1}
\end{figure*}

\section{Introduction}
With the growing popularity of short videos on social media platforms (\eg, TikTok, Instagram), video has played an increasingly important role in our daily life. However, casually captured videos are often shaky and wobbly due to amateur shooting. Although it is possible to alleviate those problems by resorting to professional equipment (\eg, dollies and steadicams), the cost of  hardware-based solutions is often expensive, making it impractical for real-world applications. By contrast, software-based or computational solutions such as video stabilization algorithms~\cite{guilluy2021video} have become an attractive alternative to improve the visual quality of shaky video by eliminating undesirable jitter. 

Existing video stabilization techniques can be classified into two categories: optimization-based and learning-based. Traditional optimization-based algorithms~\cite{liu2011subspace,morimoto1998evaluation,Bundle,Wang19,L1} have been widely studied due to their speed and robustness. 
The challenges of them are the occlusion caused by changes in depth of field and the interference caused by foreground objects on camera pose regression. Furthermore, their results often contain large missing regions at frame borders, particularly when videos with a large camera motion. 
In recent years, learning-based video stabilization algorithms~\cite{DIFRINT,FuSta,OVS,yu20} have shown their superiority by achieving higher visual quality compared to traditional methods. However, their stabilization model is too complex for rapid computation, and its generalization property is unknown due to the scarcity of training datasets.

To overcome those limitations, we present an iterative optimization-based learning approach that is efficient and robust, capable of achieving high-quality stabilization results with full-frame rendering, as shown in Fig.~\ref{fig:fig_1}. The probabilistic stabilized network addresses the issues of occlusion and interfering objects, and achieves fast pose estimation. Then the full-frame outpainting module retains the original field of view (FoV) without aggressive cropping.
An important new insight brought by our approach is that the objective of video stabilization is to suppress the implicitly embedded noise in the video frames rather than the explicit noise in the pixel intensity values. This inspired us to adopt an expectation-maximization (EM)-like approach for video stabilization.
Importantly, considering the strong redundancy of video in the temporal domain, we ingeniously consider stable video (the target of video stabilization) as the fixed point of nonlinear mapping. Such a fixed-point perspective allows us to formulate an optimization problem of the optical flow field in commonly shared regions. Unlike most methods that resort to the ad hoc video dataset~\cite{zhou2018stereo} or the deblurred dataset~\cite{su2017deep} as stabilized videos, we propose to construct a synthetic training dataset to facilitate joint optimization of model parameters in different network modules.

To solve the formulated iterative optimization problem, we take a divide-and-conquer approach by designing two modules: probabilistic stabilization network (for motion trajectory smoothing) and video outpainting network (for full-frame video rendering).
For the former, we propose to build on the previous work of PDCNet ~\cite{truong2021pdc,truong2022probabilistic} and extend it using a coarse-to-fine strategy to improve robustness. For a more robust estimate of the uncertainty of the optical flow, we infer masks from the optical flow by bidirectional propagation with a low computational cost (around $1/5$ of the time Yu~\etal~\cite{yu20}). Accordingly, we have developed a two-level (coarse-to-fine) flow smoothing strategy that first aligns adjacent frames by global affine transformation and then refines the result by warping the fields of intermediate frames. For the latter, we propose a two-stage approach (flow and image outpainting) to render full-frame video. Our experimental results have shown highly competent performance against others on three public benchmark datasets.
The main contribution of this work is threefold:
\begin{itemize}
    \vspace{-5pt}
    \item We propose a formulation of video stabilization as a fixed-point problem of the optical flow field and propose a novel procedure to generate a model-based synthetic dataset. 
    \vspace{-5pt}
    \item We construct a probabilistic stabilization network based on PDCNet and propose an effective coarse-to-fine strategy for robust and efficient smoothing of optical flow fields. 
    \vspace{-5pt}
    \item We propose a novel video outpainting network to render stabilized full-frame video by exploiting the spatial coherence in the flow field. 
\end{itemize}

\section{Related Work}
\subsection{Video Stabilization}
Most video stabilization methods can typically be summarized as a three-step procedure: motion estimation, smoothing the trajectory, and generating stable frames. Traditional methods focus primarily on 2D features or image alignment \cite{puglisi2011robust} when it comes to motion estimation. These methods are different in modeling approaches to motion, including the trajectory matrix~\cite{subspace}, epipolar geometry~\cite{epipolar,Bundle,zhao2023a2b,ye2023learning}, and the optical flow field~\cite{SteadyFlow,zhao2023learning}. Regarding the smoothing trajectory, particle filter tracking \cite{yang2009robust}, space-time path smoothing~\cite{Wang13,subspace}, and L1 optimization~\cite{L1} have been proposed. Existing methods for generating stable frames rely mainly on 2D transformations~\cite{Matsushita06}, grid warping~\cite{subspace,Bundle}, and dense flow field warping~\cite{SteadyFlow}. 

Compared to the 2D method, some approaches turn to 3D reconstruction~\cite{Liu09}. 
However, specialized hardware such as depth camera~\cite{Liu12} and light field camera~\cite{light} are necessary for these methods based on 3D reconstruction.
Some methods~\cite{adversarial,yu19,DUT,roberto2022survey} tackle video stabilization from the perspective of deep learning. 
In \cite{yu19}, the optimization of video stabilization was formulated in the CNN weight space. Recent work \cite{yu20} represents motion by flow field and attempts to learn a stable optical flow to warp frames. Another method \cite{DIFRINT} aims to learn stable frames by interpolation. These deep-learning methods generate stable videos with less distortion.

\subsection{Large FOV Video}

Unlike the early work (e.g., traffic video stabilization~\cite{ling2018stabilization}), large field of view (Field Of View) video stabilization has been attracting more researchers' attention. For most video stabilization methods, cropping is inevitable, which is why the FOV is reduced. Several approaches have been proposed to maintain a high FOV ratio. OVS~\cite{OVS} proposed to improve FOV by extrapolation. DIFRINT~\cite{DIFRINT} choose iterative interpolation to generate high-FOV frames. FuSta~\cite{FuSta} used neural rendering to synthesize high-FOV frames from feature space. 
To a great extent, the performance of interpolation-based video stabilization~\cite{DIFRINT,OVS} depends on the selected frames. If selected frames have little correspondence with each other, performance will deteriorate disastrously. Neural rendering~\cite{FuSta} synthesizes the image by weighted summing, causing blur. 
Most recently, a deep neural network~\cite{shi2022deep} (DNN) has jointly exploited sensor data and optical flow to stabilize videos.

\section{Stabilization via Iterative Optimization} \label{sec:stab_optim}

\noindent\textbf{Motivation}.
Despite rapid advances in video stabilization \cite{roberto2022survey}, existing methods still suffer from several notable limitations \cite{guilluy2021video}. First, a systematic treatment of various uncertainty factors (e.g., low-texture regions in the background and moving objects in the foreground) in the problem formulation is still lacking. These uncertainty factors often cause occlusion-related problems and interfere with the motion estimation process. Second, the high computational cost has remained a technical barrier to real-time video processing.
The motivation behind our approach is two-fold. On the one hand, we advocate for finding commonly shared regions among successive frames to address various uncertainty factors in handheld video. On the other hand, in contrast to these prestabilization algorithms~\cite{yu19,yu20,FuSta,Bundle} based on traditional approaches, we proposed a novel high-efficiency prestabilization algorithm based on probabilistic optical flow. Flow-based methods are generally more accurate in motion estimation and deserve a high time cost. Accuracy and efficiency, we have both. 


\noindent\textbf{Approach}.
We propose to formulate video stabilization as a problem of minimizing the amount of jerkiness in motion trajectories. It is enlightening to think of video stabilization as a special kind of ``video denoising'' where noise contamination is not associated with pixel intensity values, but embedded into the motion trajectories of foreground and background objects. Conceptually, for video restoration in which unknown motion is the hidden variable, we can treat video stabilization as a chicken-and-egg problem \cite{hyun2015generalized} - i.e., the objective of smoothing motion trajectories is intertwined with that of video frame interpolation. Note that an improved estimation of motion information can facilitate the task of frame interpolation and vice versa. Such an observation naturally inspires us to tackle the problem of video stabilization by iterative optimization.

\begin{figure}[!t]
  \centering
  \includegraphics[width=0.95\linewidth]{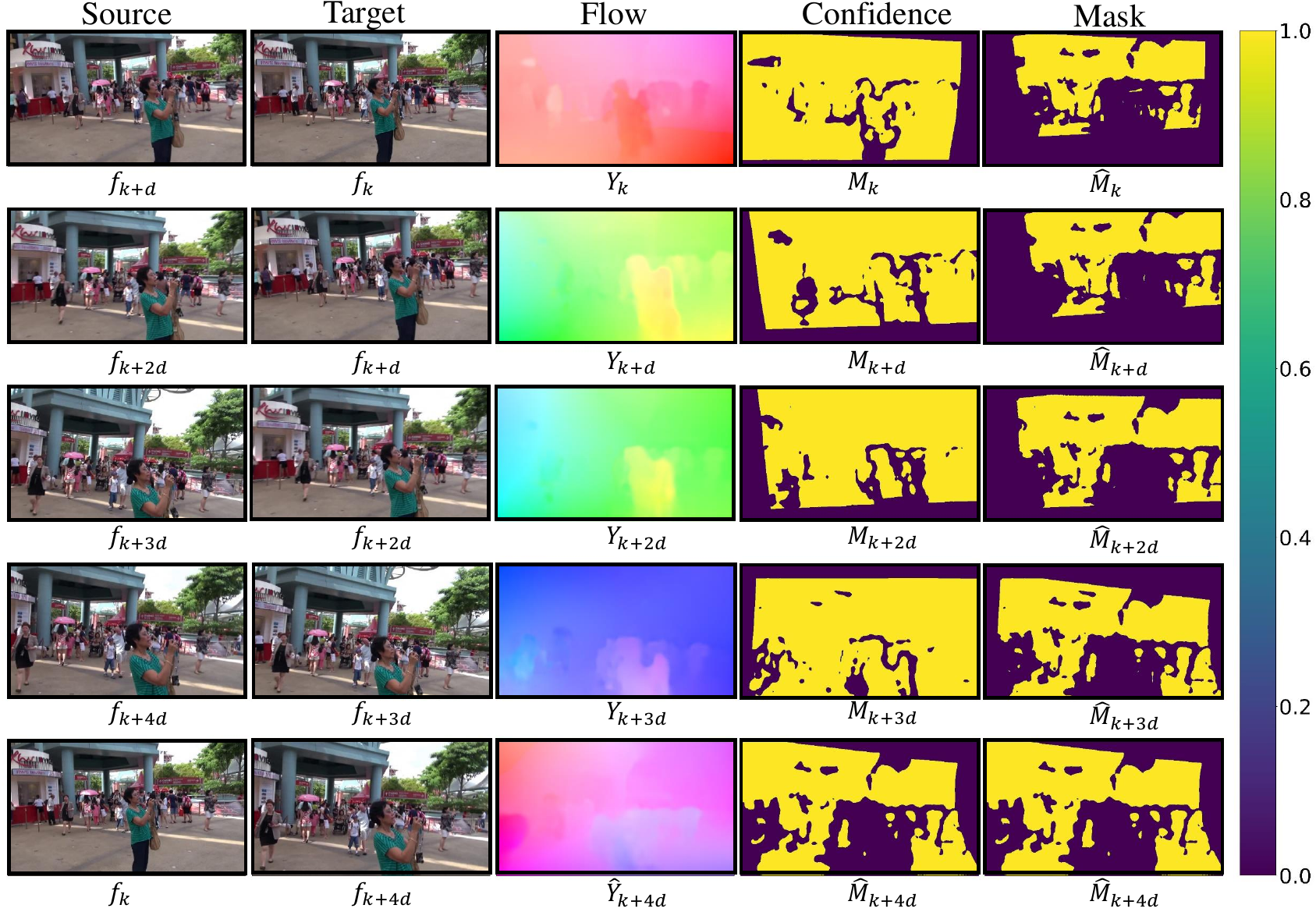}
  \caption{\textbf{Visualization of confidence map back-propagation results.} The flow field and confidence map are predicted by PDCNet between source and target images. The obtained masks in the last column represent the shared regions among these frames.} \vspace{-0.10in}
  \label{fig:coarse_process}
\end{figure}

Through divide-and-conquer, we propose to formulate video stabilization as the following optimization problem. Given a sequence of $n$ frames along with the set of optical flows $\mathcal{Y}$, we first define the confidence map, which originally indicates the reliability and accuracy of the optical flow prediction at each pixel. Here, we have thresholded the confidence map as a binary image, which represents accurate matches (as shown in the $4$-th column of Fig.~\ref{fig:coarse_process}).
Then we can denote $n$ frames and the corresponding set of confidence maps $\mathcal{M}$ by:
\begin{equation} \label{eq:two-set}
    \mathcal{Y} = \{Y_{1}, Y_{2}, \cdots,  {Y}_{q}\}\,, 
    \mathcal{M} = \{M_{1}, M_{2}, \cdots,  {M}_{q}\}\,.
\end{equation}

\section{Probabilistic Stabilization Network} 
\label{sec:video stab}

The problem of video stabilization can be formulated as finding a nonlinear mapping $f: \mathcal{Y} \rightarrow \hat{\mathcal{Y}}$ where $\hat{\mathcal{Y}}$ denotes the optical flow set of the stabilized video. We hypothesize that a desirable objective to pursue $f$ is the well-known fixed-point property, i.e., $\hat{\mathcal{Y}}=f(\hat{\mathcal{Y}})$. To achieve this objective, we aim to minimize an objective function $\mathds{F}$ characterized by the magnitude of optical flows between commonly shared regions, as represented by $\mathcal{M}$. Note that $\mathcal{M}$ is the hidden variable in our problem formulation (that is, we need to estimate $\mathcal{M}$ from $\mathcal{Y}$). A popular strategy to address this chicken-and-egg problem is to alternatively solve the two subproblems of unknown $\mathcal{M}$ and $\hat{\mathcal{Y}}$. More specifically, the optimization of $\mathds{F}$ is decomposed into the following two subproblems:

\begin{equation}
\small
\label{eq:obj}
\begin{split}
(\hat{M}_{1}, \cdots, \hat{M}_{q}) = \mathop{\arg\min}\limits_{\mathcal{M}} \mathds{F}(\hat{Y}_{1} \odot M_{1}, \cdots, \hat{Y}_{q} \odot M_{q}) \,,\\
    (\hat{Y}_{1}, \cdots, \hat{Y}_{q}) = \mathop{\arg\min}\limits_{\mathcal{Y}} \mathds{F}(Y_{1} \odot \hat{M}_{1}, \cdots, Y_{q} \odot \hat{M}_{q})\,,
    \end{split}
\end{equation}
where $Y / \hat{Y}$ denotes the magnitude of the optical flow values before and after stabilization and $\odot$ is the Hadamard product. 
Instead of analyzing the estimated motion in the frequency domain, we hypothesize that stabilized videos are the fixed points of video stabilization algorithms, which minimize the above objective function. A fixed point~\cite{agarwal2001fixed} is a mathematical object that does not change under a given transformation. Numerically, fixed-point iteration is a method of computing fixed points of a function. It has been widely applied in data science~\cite{combettes2021fixed}, and image restoration~\cite{mao2016image,chan2016plug}. Here, we denote $\mathds{F}$ in Eq.~\eqref{eq:obj} as the function to be optimized, and the fixed point of $\mathds{F}$ is defined as the stabilized video.
Next, we solve these two subproblems by constructing the module of stabilization.


\vspace{5pt}
\subsection{Probabilistic Flow Field}  \label{subsec:pose_regression_network}

First, we start with an interesting observation. When playing an unstable video at a slower speed (e.g., from 50fps to 30fps), the video tends to appear less wobbly.  It follows from the observation that the fundamental cause of video instability is the fast frequency and the large amplitude of the still objects' motion speed. Therefore, the core task of motion smoothing is to identify the region that needs to be stabilized. As Yu~\etal~\cite{yu20} pointed out, mismatches, moving objects, and inconsistent motion areas lead to variation of the estimated motion from the true motion trajectories and should be masked. \textit{An important new insight brought about by this work is that these regions of inconsistency in the optical flow fields tend to show greater variability as the frame interval increases.} Therefore, if these inconsistent regions are excluded, the video stabilization task can be simplified to the first sub-problem in Eq. \eqref{eq:obj} - i.e., minimizing $\mathds{F}$ over $\mathcal{M}$ for a fixed $\mathcal{Y}$.

To detect unreliable matches and inconsistent regions, we have adopted the probabilistic flow network -- PDCNet~\cite{truong2021pdc,truong2022probabilistic}, that jointly tackles the problem of dense correspondence and uncertainty estimation, as our building block. Suppose that PDCNet estimates the optical flow $Y_{k}$ from frame $k+d$ to frame $k$ with the resulting confidence map $C_{k}$.
Although $C_{k}$ denotes the inaccuracy of the optical flow estimate in $Y_{k}$,  it is often sensitive to the frame interval $d$. For example, it is difficult to identify the inconsistent region when $d$ is small, while the common area is less controllable when $d$ is large. Therefore, simply using an optical flow field to estimate inconsistent regions is not sufficient. 

We have designed a more robust solution to the joint estimation of dense flow and confidence map based on a coarse-to-fine strategy. The basic idea is to first obtain the probabilistic flow field at a coarse scale (e.g., with the down-sampled video sequence by a factor of $d$ along the temporal axis) and then fill in the rest (i.e., the frames between the adjacent frames in the down-sampled video) at the fine scale. Such a two-level estimation strategy effectively overcomes the limitations of PDCNet by propagating the estimation results of probabilistic flow fields bidirectionally, as we will elaborate next.

\vspace{5pt}
\noindent \textbf{Coarse-scale strategy } As shown by the second-to-last column in Fig.~\ref{fig:coarse_process}, the confidence map estimated by PDCNet can identify the mismatched region, but fails to locate the objects with small motions (\eg, people and the sky). To overcome this difficulty, we introduce the binary mask as a warped and aggregated confidence map (refer to the last column of Fig.~\ref{fig:coarse_process}). Specifically, we first propose to obtain the confidence map $\hat{C}_{k+(n-1)d}$ with down-sampled video (the last row of Fig.~\ref{fig:coarse_process}). In the forward direction, we estimate dense flow and confidence map using PDCNet; then $\hat{C}_{k+(n-1)d}$ is backpropagated to update the binary mask $\hat{M}$ by thresholding and setting intersection operators. Through bidirectional propagation, the 
region covered by $\mathcal{M}$ is the shared content from frame $k$ to frame $k+(n-1)d$.  The complete procedure can be found in Supp.~\textcolor{red}{S1}.


\vspace{5pt}
\noindent \textbf{Fine-scale strategy} Based on the coarse-scale estimation result for downsampled video (i.e., $\mathcal{M}=\{\hat{M}_{k}, \hat{M}_{k+d}, \cdots, \hat{M}_{k+(n-1)d}\}$), we fill in the missing $d-1$ frames at the fine scale. Specifically, considering the sequential frames from $k$ to $k+d$, we can obtain two sets similar to Eq.~\eqref{eq:two-set}, which are $\mathcal{Y} = \{Y_{k}, Y_{k+1}, \cdots, {Y}_{k+d-1}, \hat{Y}_{k+d}\}$ and their corresponding confidence map set $\mathcal{C}$. Note that $\hat{M}_{k+d} \in \mathcal{M}$ has been calculated in coarse stage. 
Setting $d=1$, we can call the algorithm again to obtain the set of output masks $\mathcal{M}$ for the rest of $d-1$ frames. 

\begin{figure}[!t]
  \centering
  \includegraphics[width=0.95\linewidth]{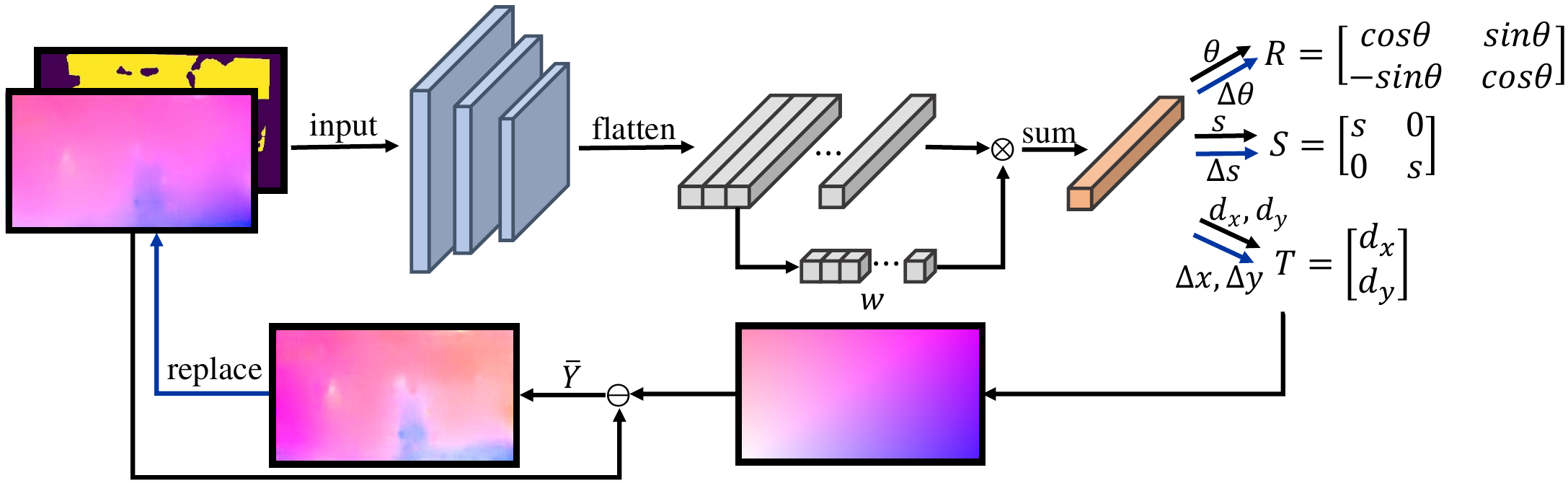}
  \caption{\textbf{Architecture of the camera pose regression network.} Given the flow field and mask, our network predicts the corresponding affine transformation parameters by closed-loop iterations.} \vspace{-0.2in}
  \label{fig:predict_atchitecture}
\end{figure}

\subsection{Coarse-scale Stabilizer} 
\label{subsec:image-level stabilizer}
To coarsely stabilize the video, we first propose aligning the adjacent frames with a global affine transformation~\cite{zhang2015global}. The optimization function $\mathds{F}$ in Eq.~\eqref{eq:obj} is represented as
\begin{equation} \label{eq:optim coarse}
    \mathbf{T}^{*} = \mathop{\arg\min}\limits_{\mathbf{T}} \mathbf{T}(Y \odot \hat{M}) \,,
\end{equation}
where $\mathbf{T}(\cdot)$ denotes the image transformation applied to the shared region $\hat{M}$ (the result of Sec.~\ref{subsec:pose_regression_network}) of the optical flow field $Y$.
Most conventional methods~\cite{DUT,yu20,FuSta} adopt image matching to obtain $\mathbf{T}(\cdot)$- \eg, keypoint detection~\cite{sift,surf,orb}, feature matching~\cite{baumberg2000reliable,sift,oanet}; and camera pose estimation is implemented by OpenCV-python. However, these methods are often time-consuming and computationally expensive. For example, two adjacent frames of unstable videos usually share a large area and are free from perspective transformation. Thus, an affine transformation, including translation, rotation, and scaling, is sufficient. More importantly, within the optimization-based learning framework, we can regress these linear parameters of $\mathbf{T}(\cdot)$ from the optical flow field, which characterize the relative coordinate transformation of the matched features.

We propose a novel camera pose regression network, as shown in Fig.~\ref{fig:predict_atchitecture}. Given an optical flow field $Y$ and the corresponding mask field $\hat{M}$, our network $\Phi(\cdot)$ can directly estimate the unknown parameters $\mathbf{T}(\cdot) \propto \{\theta,s,d_x,d_y \} = \Phi(Y, \hat{M})$. To solve the optimization problem of Eq.~\eqref{eq:optim coarse}, we use the estimated parameters to iteratively compute the corresponding residual optical flow fields such that
\begin{equation} \label{eq:refine_flow}
    \bar{Y} = Y - (S \cdot R\cdot V + T) \,,
\end{equation}
where $T$, $S$, and $R$, respectively, denote the translation, scaling, and rotation matrix, and $V \in \mathbb{R}^{2 \times H \times W}$ represents an image coordination grid. Then $\{\Delta\theta,\Delta s,\Delta x,\delta y \} = \Phi(\bar{Y}, \hat{M})$ is calculated iteratively to produce the updated parameters $\{\theta+\Delta\theta, s\cdot\Delta s, d_x +\Delta x, d_y+\Delta y\}$. The finally estimated affine transformation is smoothed by a moving Gaussian filter with a window size of $20$ pixels.

\vspace{5pt}
\noindent \textbf{Loss functions } Our loss functions include robust loss $\ell_1$ and grid loss commonly applied in consistency filtering \cite{oanet}. We directly calculate the $\ell_1$ loss between the predictions and their ground truth $\{\hat{\theta},\hat{s}, \hat{d}_x,\hat{d}_y\}$,
\begin{equation} \label{eq:L_gt}
\begin{split}
    L_{gt} =&\lambda_{\theta}\parallel\mid\theta - \hat\theta \mid\parallel_{1} + \lambda_{s}\parallel\mid 1 - \frac{s}{\hat{s}} \mid\parallel_{1} \\ \
    &+  \lambda_{t}\parallel\mid d_x - \hat{d}_x\mid + \mid d_y -\hat{d}_y \mid\parallel_{1} \,.
\end{split}
\end{equation}
For better supervision of the estimated pose, we calculate the loss $L_{grid}$ with the grid $V \in \mathbb{R}^{2\times h \times h}$ in Eq.~\eqref{eq:refine_flow}, \ie,
\begin{equation}
    L_{grid} = \parallel  (\hat{S} \cdot \hat{R} \cdot V + \hat{T}) - (S \cdot R \cdot V + T) + \epsilon  \parallel_{1}\,
\end{equation}
where $\hat{\cdot}$ denotes the ground truth and $\epsilon$ is a small value for stability. The final loss $L_{stab}$ consists of $L_{gt}$ and $L_{grid}$, 
\begin{equation} \label{eq:l_stab}
    L_{stab} = L_{gt} + \lambda_{grid} L_{grid} \,.
\end{equation}

\subsection{Fine-scale Stabilizer} \label{subsec:pixel-level stabilizer}
The assumption with an affine transformation of the coarse-scale stabilizer could cause structural discontinuity and local distortion.
Therefore, our objective is to refine the coarsely stabilized video by optical flow smoothing. Unlike Eq.~\ref{eq:optim coarse} which applies an image transformation matrix to optimize the optical flow field, we optimize it at the pixel level by a flow warping field $\mathbf{W}$. Thus, the function $\mathbb{F}$ in Eq.~\eqref{eq:obj} is given by
\begin{equation} 
    \mathbf{W}^{*} = \mathop{\arg\min}\limits_{\mathbf{W}} \sum_{i=0}^{N-1}{\mathbf{W}_i}(Y_i \odot \hat{M}_i) \,.
\end{equation}
The flow smoothing network follows the U-Net architecture in ~\cite{zhou2018stereo}. We use $N$ frames of optical flow fields $\mathbf{F}$ and mask fields $\hat{M}$ as input and obtain $(N-1)$ frame warp fields $\mathbf{W}$ of intermediate frames. Specifically, for the optical flow field $Y_{k}$ (from frame $k+1$ to frame $k$), we denote the aligned matrices of each frame in Sec.~\ref{subsec:image-level stabilizer} as $H_{k}\in \mathbb{R}^{2\times 3}$ and $H_{k+1} \in \mathbb{R}^{2\times 3}$, respectively. Then the input optical flow $\mathbf{F}_{k}\in \mathbb{R}^{2\times HW}$ can be represented by 
\begin{equation}
    \mathbf{F}_{k} = H_{k+1} \cdot [V + Y_{k} \mid 1] - H_{k}\cdot [V \mid 1] \,,
\end{equation}
where $V \in \mathbb{R}^{2 \times HW}$ and $[\cdot | 1]$ denote the normalized coordinate representation. Furthermore, to better adapt the flow smoothing network to the mask field $\hat{M}$, we fine-tune it using our synthetic dataset (as we will elaborate in Sec.~\ref{subsec:self-supervised dataset}). The loss function follows the motion loss~\cite{yu19}
\begin{equation}
    L_{smooth} = \sum_{k=0}^{N-1} (\mathbf{F}_{k} + \mathbf{W}_{k} - \mathbf{F}_{k}(\mathbf{W}_{k+1})) \odot \hat{M}\,,
\end{equation}where $\mathbf{W}_0 = \mathbf{W}_N = 0$\,.

\section{Video Outpainting Network} \label{sec:video outpainting}

\begin{figure}[!t]
 \centering
 \includegraphics[width=0.95\linewidth]{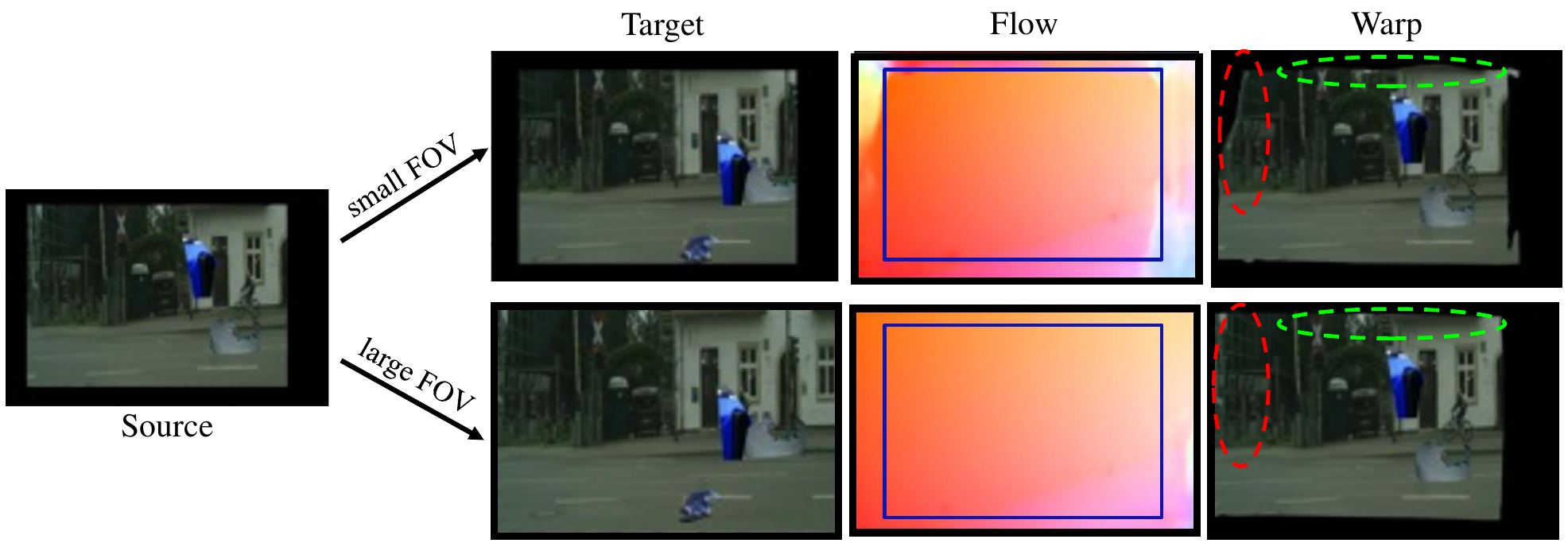}
 \caption{\textbf{Illustration of flow fields and warped images under different FOVs.} The target image with large FOV presents the smoother flow field and better warped result.} \vspace{-0.2in}
 \label{fig:flow_extrapolate}
\end{figure}

Most video stabilization methods crop an input video with a small field of view (FOV), excluding missing margin pixels due to frame warping. In contrast, full-frame video stabilization~\cite{FuSta,DIFRINT,Matsushita06} proposes to generate a video that maintains the same FOV as the input video without cropping. They directly generate stabilized video frames with large FOV by fusing the information from neighboring frames. An important limitation of existing fusion strategies is the unequal importance of different frames (i.e., the current and neighboring frames are weighted equally), which would lead to unpleasant distortions in fast-moving situations. To overcome this weakness, we propose a two-stage framework to combine flow and image outpainting strategies \cite{sabini2018painting}. In the first stage, we used flow-outpainting to iteratively align the neighboring frames with the target frame. In the second stage, we fill the target frame with adjacent aligned frame pixels by image outpainting.

\subsection{Flow Outpainting Network} \label{subsec:flow outpainting}
Let $I^{t}$ denote the outpainted target image and $I^{s}$ the neighboring source image. We aim to fill the missing pixel region $M_{\varnothing}^{t}$ of $I^{t}$ with the pixels of $I^{s}$. As shown in Fig.~\ref{fig:flow_extrapolate}, we take two different FOVs of $I^{t}$ as input and obtain the corresponding optical flow fields and warped results $I_{warp}^{s}$. The small FOV $I^{t}$ cannot guide $I^{s}$ well to fill the regions in $M_{\varnothing}^{t}$. Since the predicted optical flow field in $M_{\varnothing}^{t}$ is unreliable due to the lack of pixel guidance of $I^{t}$, the out-of-view region of $M_{\varnothing}^{t}$ has artifacts (marked in the last column of Fig.~\ref{fig:flow_extrapolate}). We observe that the optical flow field of the large FOV $I^{t}$ is continuous in $M_{\varnothing}^{t}$, which inspired us to extrapolate the flow of $M_{\varnothing}^{t}$ using the reliable flow region (the $3$rd column of Fig.~\ref{fig:flow_extrapolate}).

\begin{figure}[!t]
  \centering
  \includegraphics[width=0.96\linewidth]{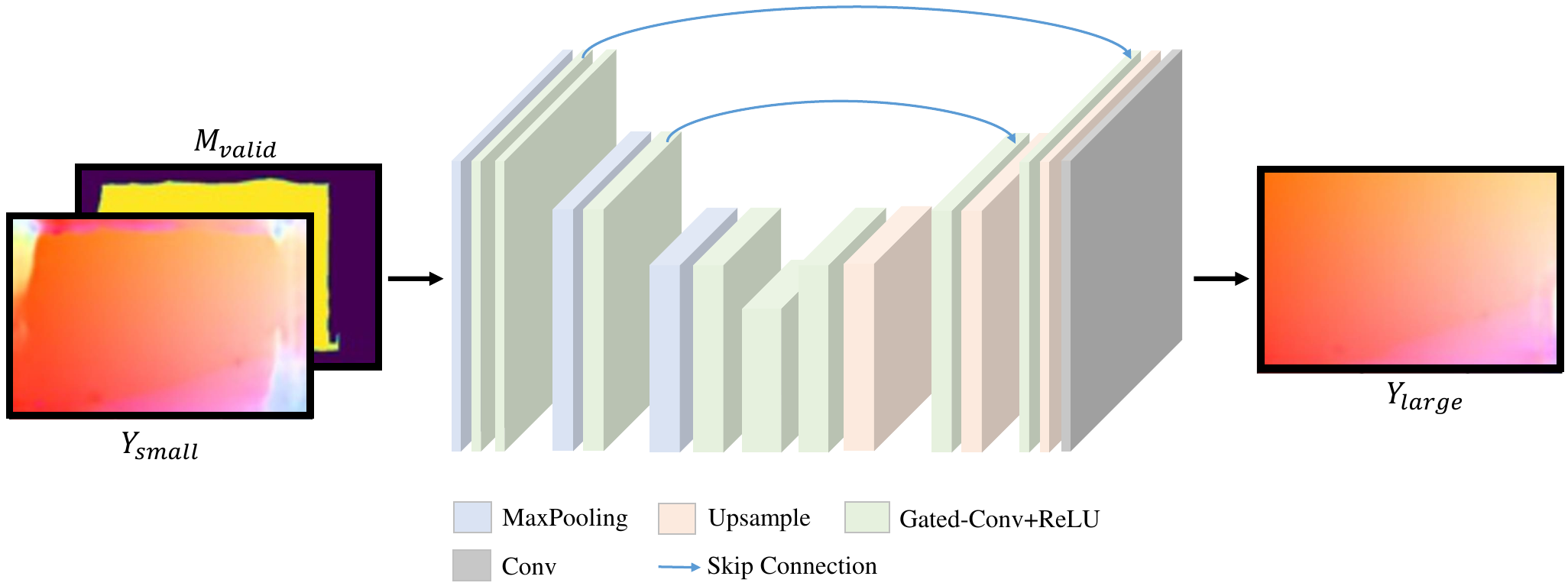}
  \caption{\textbf{Architecture of flow outpainting network.} Given a small FOV flow field and valid mask, the network predicts the large FOV flow field.} \vspace{-0.2in}
  \label{fig:outpainting_architecture}
\end{figure}

We propose a novel flow outpainting network (Fig.~\ref{fig:outpainting_architecture}), which extrapolates the large FOV flow field using a small FOV flow field and the corresponding valid mask. Specifically, we adopt a U-Net architecture and apply a sequence of gated convolution layers with downsampling / upsampling to obtain the large flow field of FOV $Y_{large}$. Note that the input flow field $Y_{small}$ and the valid mask $M_{valid}$ have been estimated by PDCNet (see Sec.~\ref{subsec:pose_regression_network}). 

\vspace{5pt}
\noindent \textbf{Loss functions } Our loss functions include robust loss $\ell_1$ and loss in the frequency domain~\cite{fourier}.  We directly calculate the $\ell_1$ loss between $Y_{large}$ and its ground truth $\hat{Y}_{large}$,
\begin{equation}
\small
\begin{split}
    L_{Y} = &\lambda_{in} \cdot \parallel \mid Y_{large} - Y_{small} \mid \odot M_{valid} \parallel_{1} \\
    &+ \lambda_{out} \cdot \parallel \mid Y_{large} - \hat{Y}_{large} \mid \odot (\sim M_{valid}) \parallel_{1} \,.
\end{split}
\end{equation}
To encourage low frequency and smoothing $Y_{large}$, we add the loss in the frequency domain $L_{F}= \parallel \hat{\mathbf{G}} \cdot \mathcal{F} Y_{large} \parallel_{2}$ \,,
where the normalized Gaussian map $\hat{\mathbf{G}}$ with $\mu = 0$ and $\sigma = 3$ is inverted by its maximum value and $\mathcal{F} Y_{large}$ denotes the Fourier spectrum of $Y_{large}$. The final loss consists of $L_{Y}$ and $L_{F}$,
\begin{equation}
    L_{outpaint} = L_{Y} + \lambda_{F}L_{F} \,.
\end{equation}

\subsection{Image Margin Outpainting}

\begin{figure}[!t]
  \centering
  \includegraphics[width=0.95\linewidth]{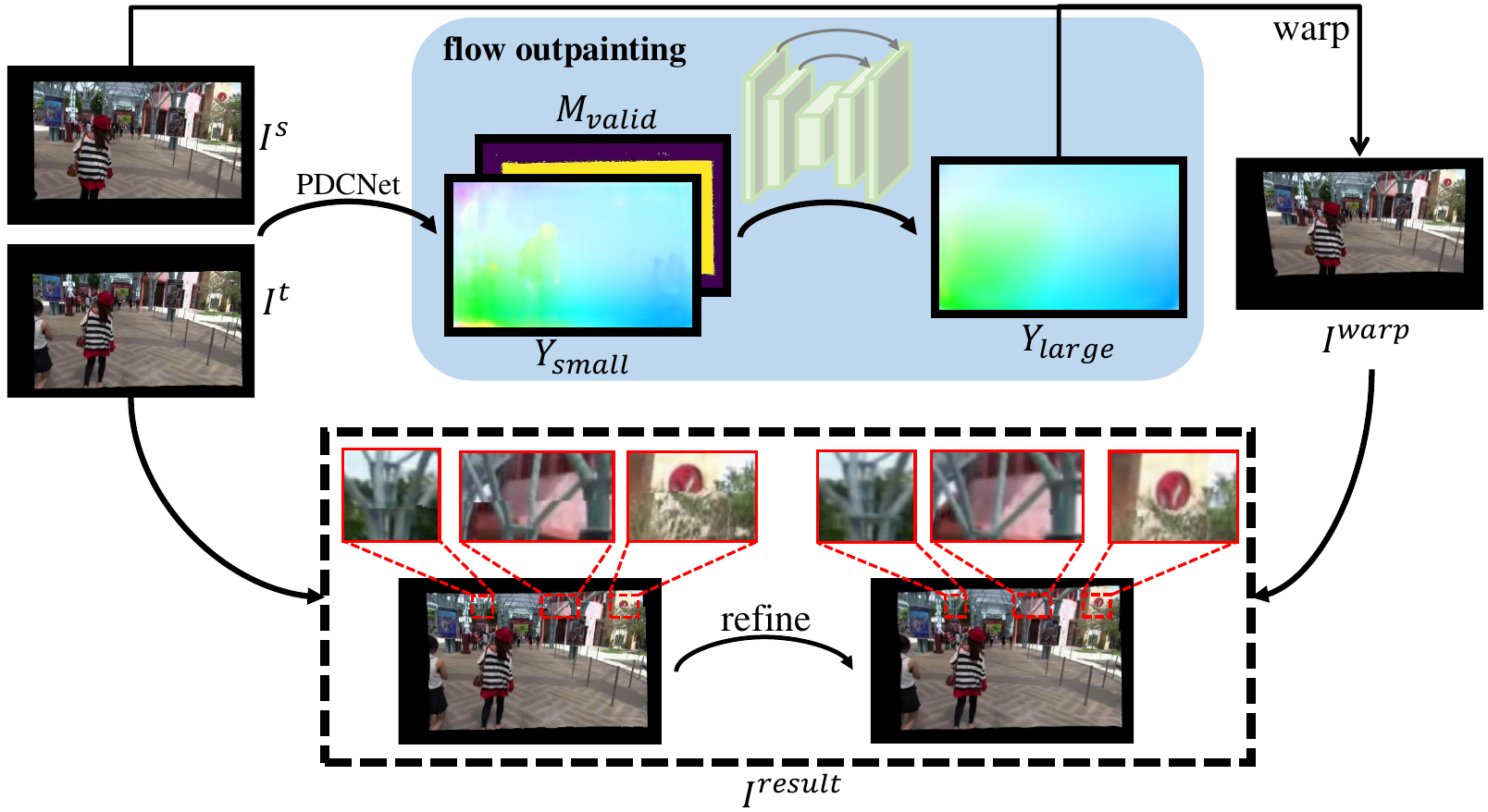}
  \caption{\textbf{Overview image margin outpainting.} Given the target frame $I^t$, the reference frame $I^s$ is coarsely aligned to $I^{warp}$ by the predicted large-FOV flow field $Y_{large}$\,. Then, we adopt a margin fusion approach to obtain the result frame $I^{result}$, by carefully aggregating $I^t$ and $I^{warp}$.}\vspace{-0.2in}
  \label{fig:image_filling}
\end{figure}

Based on our proposed flow-outpainting network, we design a margin-outpainting method by iteratively aligning frame pairs (see Fig.~\ref{fig:image_filling}). As discussed in Sec.~\ref{subsec:flow outpainting}, we can obtain a large FOV flow field $Y_{large}$. The neighboring reference frame $I^{s}$ is warped as $I^{warp} = Y_{large}(I^{s})$ as shown in Fig.~\ref{fig:image_filling}. In theory, the outpainted frame $I^{result} = I^{t}\cdot M_{valid} + I^{warp}\cdot (\sim M_{valid})$. However, we notice that there are obvious distortions at the image border. To further align the margins, we take a margin fusion approach (the detailed algorithm can be found in Supp.~\textcolor{red}{S$1$}) We crop $I^{warp}$ and $I^{t}$ to $I_{c}^{s}$ and $I_{c}^{t}$. Then, we can obtain a new warped frame $I_{c}^{warp}$ by flow outpainting. In particular, we did not choose to add $I_{c}^{warp}$ and $I^{t}$ directly. To identify the misaligned region, we propose to outpaint the mask $M_{I^t}$ by extending the watershed outward from the center. Instead of a preset threshold, we adaptively choose between the target image $I^t$ and the warped image $I_c^{warp}$. Then, the final frame $I^{result}$ consists of $I^t$ and $I_c^{warp}$: $I^{result} =  I^t \cdot M_{I^t} + I_c^{warp} \cdot (\sim M_{I^t})$ \,.
Compared to the two results $I^{result}$, our strategy successfully mitigates misalignment and distortions at the boundary of video frames.

\vspace{5pt}
\noindent \textbf{Multi-frame fusion} 
During the final stage of rendering, we use a sequence of neighboring frames to outpaint the target frame, while they may have filled duplicate regions. It is important to find out which frame and which region should be selected. We proposed the selection strategy for multi-frame fusion (the details can be found in Supp.~\textcolor{red}{S1}). By weighing the metric parameters of each frame, we finally obtain the target frame with large FOV. Note that each frame has an added margin in the stabilization process, so we need to crop them to the original resolution. Although we have outpainted the target frame, some missing pixel holes may still exit at boundaries. Here, we apply the state-of-the-art LaMa image inpainting method~\cite{lama} to fill these holes using nearest-neighbor interpolation.


\section{Experiments} \label{sec:Experiments}

\begin{table*} \small
    \centering
    \renewcommand{\arraystretch}{1.0}
    \addtolength{\tabcolsep}{1.5pt}
    \begin{tabular}{@{}lccccccccc@{}}
\toprule
  \multirow{2}{*}{Method}  & \multicolumn{3}{c}{NUS dataset~\cite{Bundle}} & \multicolumn{3}{c}{DeepStab dataset~\cite{Wang19}}  & \multicolumn{3}{c}{Selfie dataset~\cite{yu2018selfie}} \\ \cmidrule(lr){2-4} \cmidrule(lr){5-7} \cmidrule(lr){8-10}
  ~ &C.$\uparrow$ &D.$\uparrow$ &S.$\uparrow$ &C.$\uparrow$ &D.$\uparrow$ &S.$\uparrow$ &C.$\uparrow$ &D.$\uparrow$ &S.$\uparrow$\\
\midrule
Grundmann~\etal~\cite{L1}    &0.71  &0.76 &0.62  &0.77 &0.87 &0.80 &0.75 &0.81 &0.83\\
Liu~\etal~\cite{Bundle}      &0.81  &0.78 &0.82 &0.80  &0.90  &\textbf{0.85} &0.74 &\textbf{0.89} &0.8\\
Wang~\etal~\cite{Wang19}     &0.67  &0.72 &0.41 &-  &-  &- &0.68 &0.71 &0.82\\
Yu and Ramamoorthi~\cite{yu19}  &0.78  &0.77 &0.82 &0.85 &0.89 &0.76 &0.79 &0.77 &0.84\\
Yu and Ramamoorthi~\cite{yu20} &{0.85}&0.81  &\textbf{0.86} &\underline{0.87}  &\underline{0.92} &0.82 &\underline{0.83} &\underline{0.87} &\underline{0.86}\\
Yu~\cite{yu20}+OVS$^{*}$~\cite{OVS}  &\underline{0.92} &0.78 &0.83 &-  &-  &- &-  &-  &-\\
DUT~\cite{DUT} &0.71 &0.81 &0.83 &-  &-  &- &-  &-  &-\\
DIFRINT~\cite{DIFRINT}  &\textbf{1.00}  &0.85 &\underline{0.84}  &\textbf{1.00} &0.91 &0.78 &\textbf{1.00} &0.78 &0.84\\
FuSta~\cite{FuSta}    &\textbf{1.00}  &\underline{0.87} &\textbf{0.86} &\textbf{1.00} &\underline{0.92} &0.82 &\textbf{1.00} &0.83 &\textbf{0.87}\\
Ours  &\textbf{1.00} &\textbf{0.91} &\textbf{0.86} &\textbf{1.00} &\textbf{0.94} &\underline{0.84} &\textbf{1.00}         &\underline{0.87}    &\textbf{0.87}\\
    \bottomrule
    \end{tabular}
    \caption{Quantitative results on the NUS dataset~\cite{Bundle}, the DeepStab dataset~\cite{Wang19} and the Selfie Dataset~\cite{yu2018selfie}. We evaluate the following metrics: Cropping Ratio(C.), Distortion Value(D.), Stability Score(S.). $*$ indicates the results obtained from original paper. We highlight the best method in \textbf{bold} and {\underline{underline}} the second-best.}
    \label{tab:average_tab}
    \vspace{-0.2in}
\end{table*}

\subsection{Synthetic Datasets for Supervised Learning} \label{subsec:self-supervised dataset}

Due to the limited amount of paired training data, we propose a novel model-based data generation method by carefully designing synthetic datasets for video stabilization. For our base synthetic dataset, we used a collection of images from the DPED~\cite{ignatov2017dslr}, CityScapes~\cite{cordts2016cityscapes} and ADE-20K~\cite{zhou2019semantic} datasets. 
To generate a stable video, we randomly generate the homography parameters for each image, including rotation angle $\theta$, scaling $s$, translations $(d_x,d_y)$ and perspective factors $(p_x, p_y)$. Then we divide these transformation parameters into $N$ bins equally and obtain a video of $N$ frames by homography transformations. To simulate the presence of moving objects in real scenarios, the stable video is further augmented with additional independently moving random objects. To do so, the objects are sampled from the COCO dataset~\cite{coco} and inserted on top of the synthetic video frames using their segmentation masks. Specifically, we randomly choose $m$ objects (no more than $5$), and generate randomly affine transformation parameters for each independent of the background transformation. Finally, we cropped each frame to $720 \times 480$ around its center.
For different training requirements, we apply various combinations of synthetic dataset. The implementation and training details can be found in Supp.~\textcolor{red}{S2}\,,~\textcolor{red}{S3}.

\subsection{Quantitative Evaluation} \label{subsec:quantitative}
We compare the results of our method with various video stabilization methods, including Grundmann~\etal~\cite{L1}, Liu~\etal~\cite{Bundle}, Wang~\etal~\cite{Wang19}, Yu and Ramamoorthi~\cite{yu19,yu20}, DUT~\cite{DUT}, OVS~\cite{OVS}, DIFRINT~\cite{DIFRINT}, and FuSta~\cite{FuSta}. We obtain the results of the compared methods from the videos released by the authors or generated from the publicly available official implementation with default parameters or pre-trained models. Note that OVS~\cite{OVS} does not honor their promise to provide code, thus we only report the results from their paper.
 
\vspace{5pt}
\noindent\textbf{Datasets.}
We evaluate all approaches on the NUS dataset~\cite{Bundle}, DeepStab dataset~\cite{Wang19}, and Selfie dataset~\cite{yu2018selfie}. The NUS dataset consists of $144$ videos and the corresponding ground truths in $6$ scenes. The DeepStab dataset contains $61$ videos and the Selfie dataset consistsof $33$ videos.

\vspace{5pt}
\noindent\textbf{Metrics.}
We introduce three metrics widely used in many methods~\cite{yu20,FuSta,DIFRINT} to evaluate our model:\textbf{1)}~\emph{Cropping ratio} measures the remaining frame area after cropping off the invalid boundaries. \textbf{2)}~\emph{Distortion value} evaluates the anisotropic scaling of the homography between the input and output frames. \textbf{3)}~\emph{Stability score} measures the stability of the output video. We calculate the metrics using the evaluation code provided by DIFRINT.

\vspace{5pt}
\noindent\textbf{Quantitative comparison.}
The results of the NUS dataset~\cite{Bundle} are summarized in Table~\ref{tab:average_tab} (Per-category result can be found in Supp.~\textcolor{red}{S5}). Overall, our method achieves the best distortion value compared to the state-of-the-art method, FuSta~\cite{FuSta}. Especially in the quick-rotation and zoom categories, our method outperforms pure image generation methods~\cite{FuSta,DIFRINT}. We suspect that the reduction of the shared information between frames causes the image generation methods to prefer artifacts. However, our method can ensure local structural integrity when outpainting the margin region. Furthermore, our method achieves an average cropping ratio of $1.0$ and stability scores comparable to recent approaches~\cite{DIFRINT,FuSta,yu19,yu20}. Since FuSta~\cite{FuSta} uses Yu~\etal~\cite{yu20} to obtain stabilized input videos, they have the same stability scores. It is important to note that although the stability scores are competitive, our method runs $5$ times faster than ~\cite{yu20} in the video stabilization stage. Moreover, the comparison results on DeepStab dataset~\cite{Wang19} and Selfie dataset~\cite{yu2018selfie} are also reported in Table~\ref{tab:average_tab}. Our method still shows effectiveness in different datasets, proving the generalizability of the proposed method. Note that, due to that Wang~\etal~\cite{Wang19} is trained on the DeepStab dataset, we do not report its results on the DeepStab dataset for a fair comparison. Qualitative comparisons can be found in Supp.~\textcolor{red}{S4} and \textit{supplementary video}.

\subsection{Ablation Study} \label{subsec:ablation study}

\begin{table}[!t] \small
    \renewcommand{\arraystretch}{1.0}
    \addtolength{\tabcolsep}{1.2pt}
    \centering
    \begin{tabular}{@{}lccc@{}}
\toprule
    Stabilizer &~ &Stability$\uparrow$ &Distortion$\uparrow$ \\
    \midrule
    \multirow{2}{*}{Image-level} &w.o. mask  &0.76  &0.88\\
                ~           &with mask  &0.83   &0.91 \\
    \midrule
    \multirow{2}{*}{Pixel-level} &w.o. mask  &0.81   &0.84\\
                ~           &with mask  &0.83   &0.91 \\

    \bottomrule
    \end{tabular} 
    \caption{The importance of mask to video stabilization. We validate on the Crowd category of NUS dataset~\cite{Bundle}.}
    \label{tab:ablation 1}
    \vspace{-0.20in}
\end{table}

\noindent\textbf{Importance of mask generation}. We investigate the influence of the mask on video stabilization at different stages. To better demonstrate the necessity of mask in complex scenarios, we choose the Crowd category of NUS dataset~\cite{Bundle} which includes a display of moving pedestrians and occlusions. Stability and distortion at different settings are shown in Table~\ref{tab:ablation 1}. It can be seen that the performance with mask increases significantly in both stabilizers. Specifically, the mask can both improve stability globally and alleviate image warping distortion locally. This result demonstrates the importance of mask generation for video stabilization.

\vspace{5pt}
\noindent\textbf{Flow outpainting}. We compare our flow outpainting method with the traditional flow inpainting method PCAFlow~\cite{PCAFlow}. Following~\cite{yu20} we fit the first $5$ principal components proposed by PCAFlow to the $M_{valid}=1$ regions of the optical flow field, and then outpaint the flow vectors in the $M_{valid}=0$ regions with reasonable values of the PCA Flow fitted. The result is obtained by warping the source image with the outpainted optical flow field. We perform this comparison on our synthetic validation set and evaluate it with the corresponding ground truth. Additionally, we use PSNR and SSIM~\cite{ssim} to evaluate the quality of the results. As shown in Table~\ref{tab:ablation 2}, ours dramatically outperforms PCAFlow~\cite{PCAFlow} in all objective metrics.

\vspace{5pt}
\noindent\textbf{Importance of image filling strategies}. We explore the following proposed strategies for image filling: margin outpainting, mask generation $M_{I^t}$, and multiframe selection. We isolate them from our method to compare their results with the complete version. The results are shown in Table~\ref{tab:ablation 2}. The proposed strategies are generally helpful in improving image quality. Especially, margin outpainting and mask $M_{I^t}$ are crucial to the results.

\begin{table} \small
    \renewcommand{\arraystretch}{0.9}
    \addtolength{\tabcolsep}{0.6pt}
    \centering
    \begin{tabular}{@{}lccc@{}}
\toprule
    ~ &PSNR$\uparrow$ &SSIM$\uparrow$ &Distortion$\uparrow$ \\
    \midrule
    PCAFlow~\cite{PCAFlow}    &11.34  &0.50   &0.72\\
    Our flow outpainting  &19.65    &0.77   &0.93 \\
    \midrule
    w.o. margin outpainting  &18.04 &0.64 &0.81\\
    w.o. mask $M_{I^t}$ &20.35 &0.78 &0.86\\
    w.o. multi-frame selection  &21.99 &0.84 &0.90\\
    Ours &23.17 &0.86 &0.91\\

    \bottomrule
    \end{tabular} 
    \caption{Ablation study of image filling. The evaluation is conducted on our model-based synthetic validation set.}
    \label{tab:ablation 2}
    \vspace{-0.1in}
\end{table}

\subsection{Runtime Comparison} 
\label{subsec:runtime}

\begin{table} \small
    \renewcommand{\arraystretch}{1.0}
    \addtolength{\tabcolsep}{1.5pt}
    \centering
    \begin{tabular}{@{}lr@{}}
        \toprule
        Method  &Runtime\\
        \midrule
        Traditional pose regression &216ms \\
        Our pose regression network &21ms \\
        \midrule
        Grundmann~\etal~\cite{L1}     &480ms\\
        Liu~\etal~\cite{Bundle}           &1360ms\\
        Wang~\etal~\cite{Wang19}          &460ms \\
        Yu and Ramamoorthi~\cite{yu19}  &1610ms\\
        Yu and Ramamoorthi~\cite{yu20}  &570ms\\
        DIFRINT~\cite{DIFRINT}  &1530ms\\
        FuSta~\cite{FuSta}  &9820ms\\
        Ours                        &97ms\\
    \bottomrule
    \end{tabular}
    \caption{Per-frame runtime comparison of camera pose regression and video stabilization.}
    \label{tab:runtime}
    \vspace{-0.20in}
\end{table}

Our network and pipeline are implemented with PyTorch and Python. Table~\ref{tab:runtime} is a summary of the runtime per frame for various competing methods. All timing results are obtained on a desktop with an RTX3090Ti GPU and an Intel(R) Xeon(R) Gold 6226R CPU. First, we compare the run-time of pose regression. Traditional pose regression is time consuming in feature matching~\cite{sift,ransac}, homography estimation, and SVD decomposition~\cite{svd}. Although our learning-based pose regression network runs $10$ times faster than the traditional framework. Then, we report the average run time of different methods, including optimization-based~\cite{Bundle,yu19} and learning-based~\cite{L1,Wang19,yu20,DIFRINT,FuSta}. Our method takes $97$ms which gives $\sim 5$x speed-up. This is because our method computes the optical flow field {\em only once} and without the help of other upstream task methods and manual optimization.

\subsection{Fixed-point Experiment} \label{subsec:fixed_point}
\begin{figure}[!t]
\centering
    \subfigure[shaky frames]{
    \label{fig:fixed point a}
    \centering
    \includegraphics[width=0.23\textwidth]{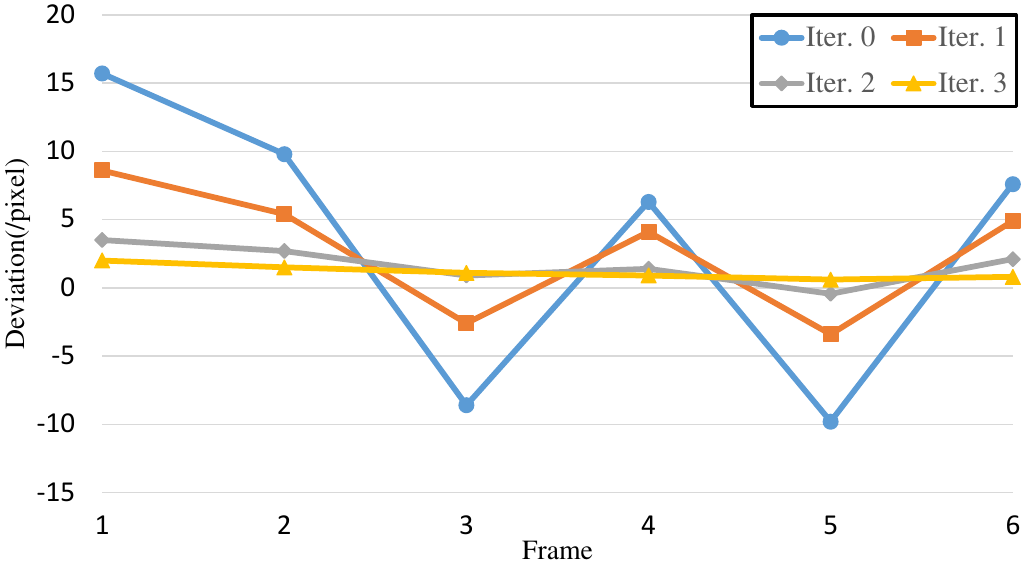}}
    \subfigure[stable frames]{
    \label{fig:fixed point b}
    \centering
    \includegraphics[width=0.22\textwidth]{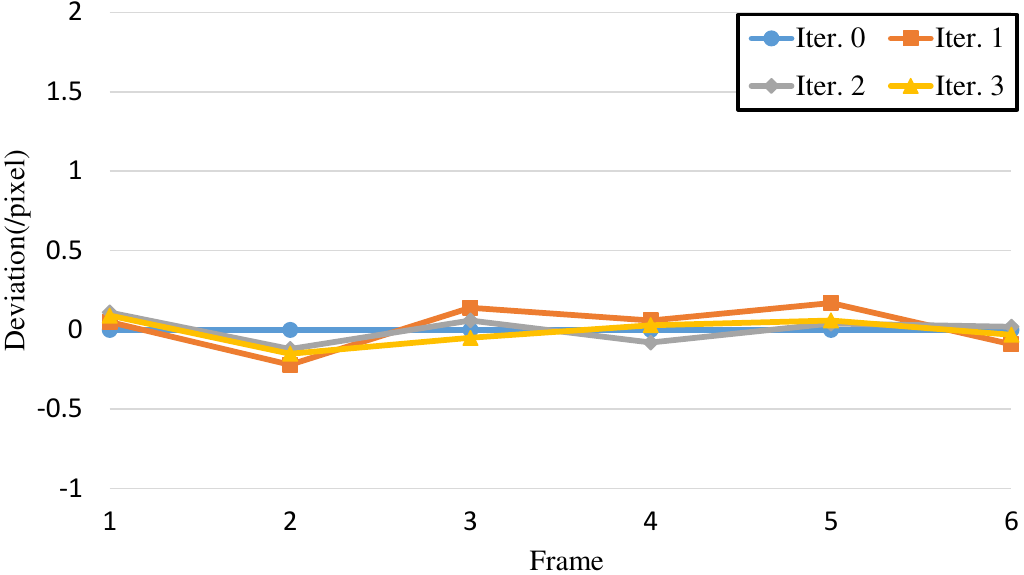}}
    \caption{\textbf{Experiment of fixed-point iteration on shaky and stable frames.} (a) Our fixed-point iteration helps the shaky frames converge to a steady state. (b) For stable frames, the fixed-point iteration guarantees the stability of results.
    } \vspace{-0.2in}
\end{figure}

To demonstrate the stability of our fixed-point optimization solution, we performed an interesting toy experiment. 
We input a sequence of shaky frames into our coarse-to-fine stabilizers, and the stabilized result will be iteratively re-stabilized by our stabilizers. For each iteration, we calculate the average magnitude of optical flow filed with global transformation and flow warping for each frame. Specifically, the regions where we calculate are marked by $\hat{M}$. As shown in Fig.~\ref{fig:fixed point a}, the deviation of the shaky frames decreases rapidly with each iteration.
Furthermore, we have pointed out that plugging a stabilized video into the stabilization system should not have an impact on the input. Thus, we plug stable frames 
into the coarse-to-fine stabilizers and iteratively stabilize them. The result is shown in Fig.~\ref{fig:fixed point b}. The deviation of each frame is perturbed around the value of zero. Obviously, our method has no effect on stable frames, which shows that stabilized video is indeed the fixed point of our developed stabilization system.

\section{Conclusion}

In this paper, we have presented a fast full-frame video stabilization technique based on the iterative optimization strategy. Our approach can be interpreted as the combination of probabilistic stabilization network (coarse-to-fine extension of PDC-Net) and video outpainting outwork (flow-based image outpainting). When trained on synthetic data constructed within the optimization-based learning framework, our method achieves state-of-the-art performance at a fraction of the computational cost of other competing techniques. It is also empirically verified that stabilized video is the fixed point of the stabilization network.

{\small
\bibliographystyle{ieee_fullname}
\bibliography{egbib}
}

\clearpage
\appendix
\begin{appendices}

In this supplementary, we will expand more details that are not included in the main text due to the page limitation. 

\section{Algorithm Details}
We introduce $3$ algorithms in the main paper. In this section, we supplement the algorithm details of the confidence map back-propagation, margin fusion approach and multi-frame fusion strategy in the main paper.

\vspace{5pt}
\noindent \textbf{Confidence map back-propagation.} Algorithm~\ref{alg:recursive} summaries the strategy of confidence map back-propagation in the main paper Section~4.1. The parameters in Algorithm~\ref{alg:recursive} are set to $k=5$, $d=10$, and $\delta_{C}=0.5$.

\begin{algorithm}[!h]
  \renewcommand{\algorithmicrequire}{\textbf{Input:}}
 \renewcommand{\algorithmicensure}{\textbf{Output:}}
  \caption{Back-propagation for aggregated confidence map }
  \label{alg:recursive}
  \begin{algorithmic}[1]  
    \Require
      $\mathcal{Y}$: optical flow;
      $\mathcal{C}$: confidence map;
      $\delta_{C}$: threshold for confidence map;
      $d$: sampling interval;
      $k$: index of the first frame;
    \Ensure
      $\mathcal{M}$: updated mask containing aggregated confidence map;
      \State set $m_{pre} = \mathds{1}(\hat{C}_{k+(n-1)d} - \delta_{C}) \in \mathcal{C}$, put $m_{pre}$ into $\mathcal{M}$;
      \For{$i=n-1$; $i>=0$; $i--$ }
            \State the optical flow field $Y_{warp} = Y_{k+id} \in \mathcal{Y}$;
            \State using $Y_{warp}$ to warp $m_{pre}$ to $\hat{m}_{pre}=Y_{warp}(m_{pre})$;
            \State the binarized confidence map $m_{new} = \mathds{1}(C_{k+id} - \delta_{C})$, where $M_{k+id} \in \mathcal{C}$;
            \State the final mask field $\hat{M}_{k+id} = \hat{m}_{pre}\,\&\, m_{new}$
            \State put $\hat{M}_{k+id}$ into $\mathcal{M}$;
            \State $m_{pre} = \hat{M}_{k+id}$
      \EndFor
  \end{algorithmic}
\end{algorithm}

\begin{figure*}[!t]
  \centering
  \includegraphics[width=0.95\linewidth]{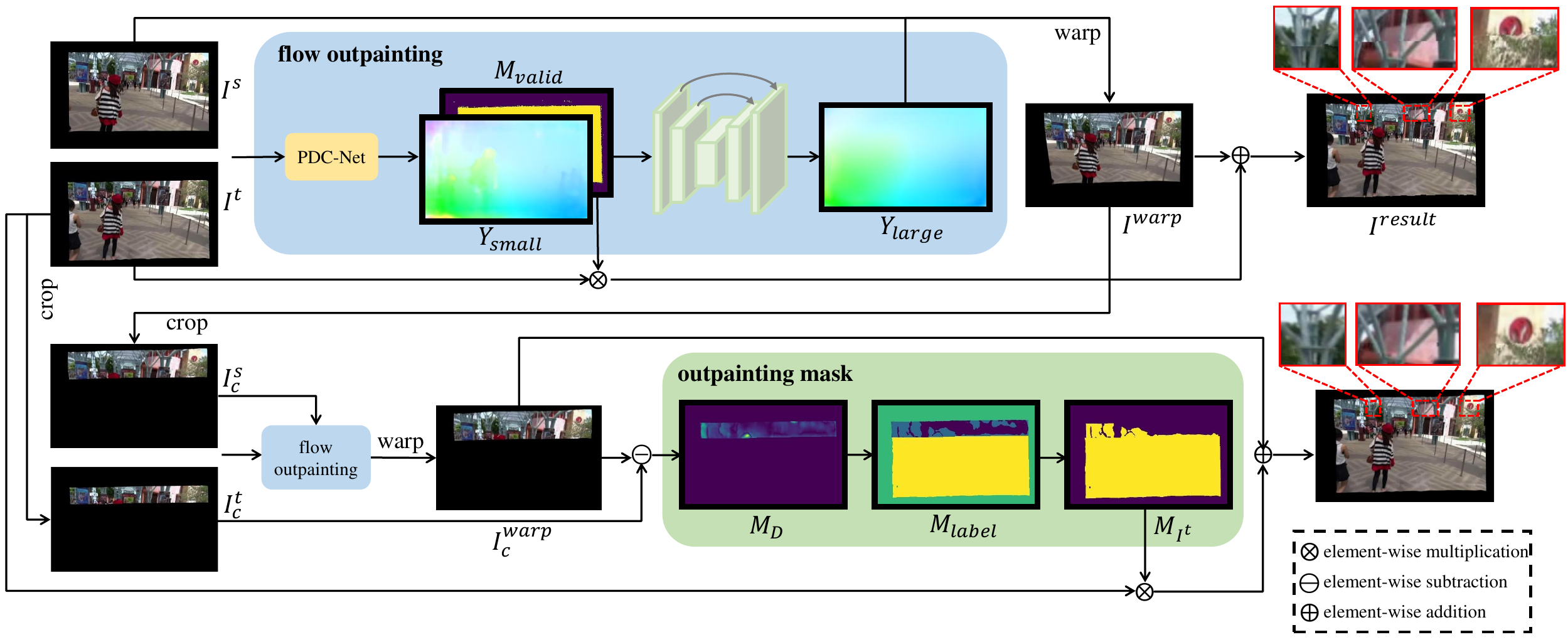}
  \caption{\textbf{Pipeline of margin fusion approach.} Given the target frame $I^t$, the reference frame $I^s$ is coarsely aligned to $I^{warp}$ by the predicted large-FOV flow field $Y_{large}$\,. Then, $I^t$ and $I^{warp}$ are cropped and re-aligned. Per Algorithm~\ref{alg:mask}, the deduced mask $M_{I^t}$ is fused with $I^t$ and $I_c^{warp}$ to obtain the resulting frame. }
  \label{fig:pipe_image_filling}
\end{figure*}

\vspace{5pt}
\noindent{\textbf{Margin fusion}}. The complete pipeline of the margin fusion approach is shown in Fig.~\ref{fig:pipe_image_filling}. At first, we coarsely align the reference frame $I^s$ and the target frame $I^t$.  We then crop $I^{warp}$ and $I^{t}$ to $I_{c}^{s}$ and $I_{c}^{t}$\,, and re-align them by the optical flow outpainting. Per Algorithm~\ref{alg:mask}, we further calculate the mask $M_{I^t}$ which indicates the chosen regions of $I^t$. The final result $I^{result}$ is obtained by combining $I^t$, $M_{I^t}$, and $I_c^{warp}$\,. The parameters in the Algorithm~\ref{alg:mask} are set to $\delta_{D}=0.2$, $\eta_{t}=20$, and $k_{t_{in}}=11$.

\begin{algorithm}[!t]
  \renewcommand{\algorithmicrequire}{\textbf{Input:}}
 \renewcommand{\algorithmicensure}{\textbf{Output:}}
  \caption{Outpainting mask Algorithm}
  \label{alg:mask}
  \begin{algorithmic}[1]  
    \Require
      $I^t$: target frame;
      $I_c^t$: cropped target frame;
      $I_c^s$: cropped source frame;
      $I_{c}^{warp}$: warped frame of $I_c^s$;
      $M^t$: valid mask of $I^t$;
      $M_c^t$: valid mask of $I_c^t$;
    \Ensure
       $M_{I^t}$: unchanged mask of $I^t$;
       \State extract feature maps with VGG-16 network $f_c^t = VGG(I_c^t) $\,, $f_c^{warp} = VGG(I_c^{warp})$;
       \State calculate the Euclidean distance in feature space $D=\parallel f_c^t - f_c^{warp} \parallel_{2}$;
       \State $M_{D} = D < \delta_{D}$; 
       \State labeled region $M_{label}$;
       \For{$i,j=0$; $i<h,j<w$; $i++,j++$ }
       \If{$\sim M^t[i,j]$} $M_{label}[i,j]=1$;
       \ElsIf{$M^t[i,j] \& (\sim M_c^t[i,j])$} $M_{label}[i,j]=2$;
       \ElsIf{$M^t[i,j] \& M_c^t[i,j] \& M_D[i,j]$} $M_{label}[i,j]=0$;
       \Else{$M_{label}[i,j]=-1$};
       \EndIf
       \EndFor
        
       \State $t_{in}=M_{label}, t_{out}=0, flag=True$;
       \While{{\tt{Sum}}$(t_{in}-t_{out})> \eta_{t}$} 
       \If{flag} 
            \State $t_{in} = t_{out}, flag=False$;
       \EndIf
       \State inflate $t_{in}$ with kernel size $k_{t_{in}}$, obtain $t_{out} = {\tt{inflate}}(t_{in})$; 
       \State $t_{out}[M_{label} == 1] = 1$;
       \State $t_{out}[M_{label} == -1] = -1$;
       \EndWhile
        
       \State $M_{I^t}= (t_{out}==2)$;
       
  \end{algorithmic}
\end{algorithm}

\vspace{5pt}
\noindent{\textbf{Multi-frame fusion}}. To adaptively determine which frame and which region should be selected, the multi-frame fusion strategy is illustrated in Algorithm~\ref{alg:selection}. The parameters in the Algorithm~\ref{alg:selection} are set to $\eta_{u}=25k$, $\eta_{r}=1.2$, and $\eta_{s}=2k$.

\begin{algorithm}[!t]
  \renewcommand{\algorithmicrequire}{\textbf{Input:}}
 \renewcommand{\algorithmicensure}{\textbf{Output:}}
  \caption{Multi-frame Fusion Algorithm}
  \label{alg:selection}
  \begin{algorithmic}[1]  
    \Require
      $I^t$: target frame;
      $I_{c_k}^{warp}$: warped of cropping source frame $I_k^s$;
      $I_{k}^{result}$: margin outpainting result of $I_{c_k}^{warp}$;
      $M_{k}^{warp}$: valid mask of $I_{c_k}^{warp}$;
    \Ensure
       $I^{fuse}$: output fusion frame;
    
    \State calculate the filling area $A_k^{s}$, misaligned region area $A_k^{u}$, and corresponding IoU ration $S_k = A_k^{u} / (A_k^{s}+1)$ of $I_{c_k}^{warp}$;
    \State sorted by $A_k^{s}$ to obtain index list $IDs$;
    \State $I^{fuse}=I^t, M^{fuse}=M_{k}^{warp}$;
    \For{k in $IDs$}
        \If{$(A_k^{u} < \eta_{u}) \& (S_k > \eta_{r}) \& (A_k^{s} > \eta_{s})$} 
        \State compute overlapped area $A_k^{o}$ between $I_{c_k}^{warp}$ and $I^{fuse}$;
        \If{$(A_k^{o}/A_k^{s} < \delta_{r})$}
            \State $I^{fuse} = I^{fuse} \cdot (\sim M_{k}^{warp}) + I_{k}^{result} \cdot M_{k}^{warp}$
        \EndIf
        \Else \State ${\tt{continue}}$;
        \EndIf
    \EndFor
  \end{algorithmic}
\end{algorithm}

\section{Synthetic Dataset for Training}

\begin{figure*}[!h]
  \centering
  \includegraphics[width=0.95\linewidth]{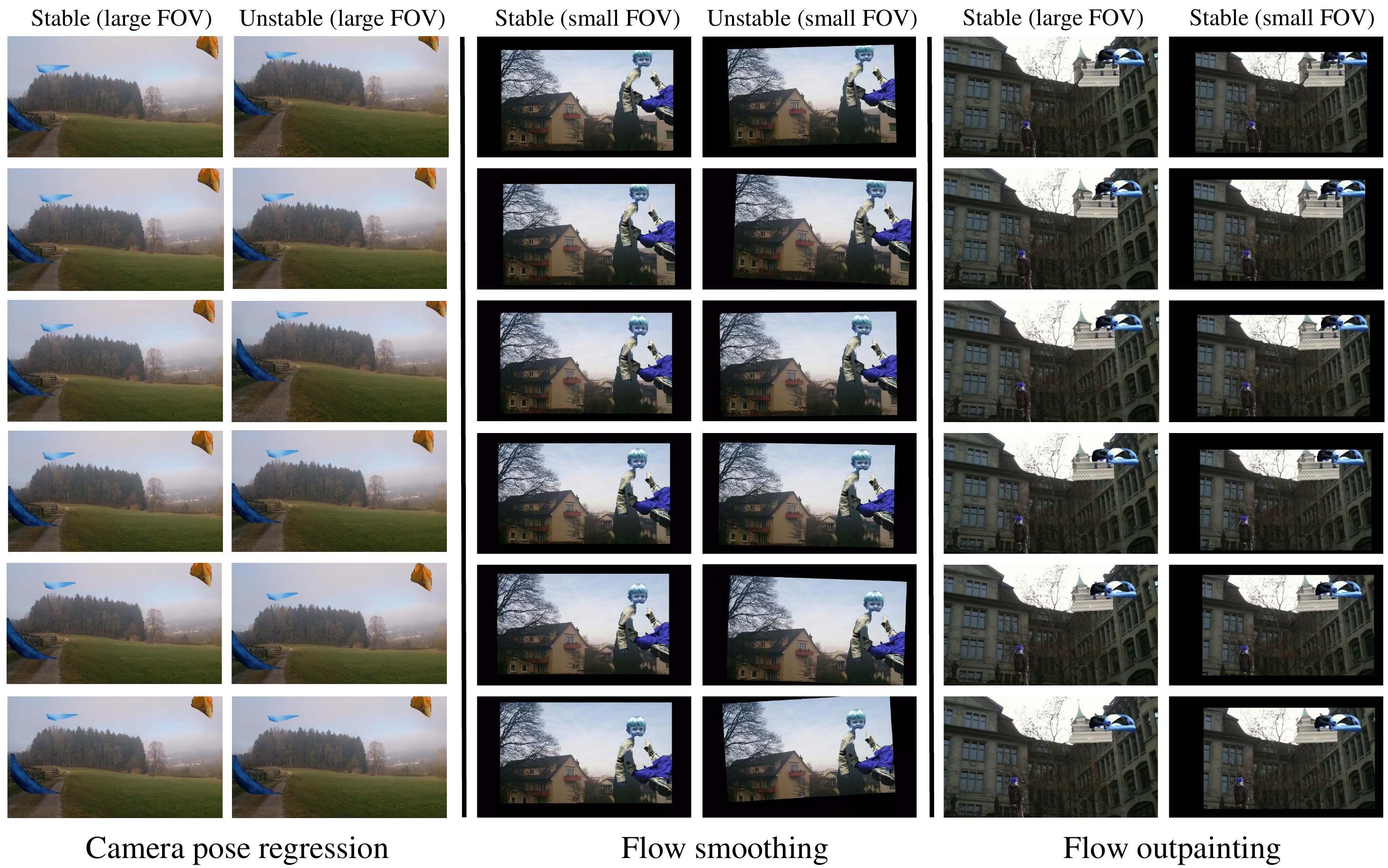}
  \caption{\textbf{Visualization of our model-based synthetic dataset.} We designed different combinations of dataset for varying tasks.}
  \label{fig:vis_dataset}
\end{figure*}

We proposed a model-based synthetic dataset in this paper. The settings of the homography parameters are as follows:
The maximum rotation angle $\theta$ is set to $10^{\circ}$. The range of scaling $s$ is set to $0.7 \sim 1.3$. The maximum translations $(d_x,d_y)$ in the $x$ and $y$ directions are $100$ and $70$, respectively. The maximum perspective factors in the $x$ direction and in the $y$ direction are $0.1$ and $0.15$.

For different training requirements, we apply various combinations of synthetic dataset, as shown in Fig.~\ref{fig:vis_dataset} (more visualizations can be found in the \textcolor{red}{\textit{Supplementary Video}}). For camera pose regression, we use the large FOV video pair of stable and unstable. For training the flow smoothing network, we alternatively adopt small FOV video pairs, which simulate coarsely stabilized video. Aiming at the flow outpainting network, we take small-FOV stable videos for training and large-FOV for ground-truth supervising.

\vspace{5pt}
\noindent\textbf{Data for Camera Pose Regression}.
For training the camera pose regression network, we need to generate unstable videos. For every frame, a random homography matrix produces an unstable frame. In practice, the perspective effects in the $x$ direction and the $y$ direction are restricted to $1e^{-5} \sim 5e^{-5}$. The pose between two unstable frames is parameterized by rotation, scaling, and translation. 

\noindent\textbf{Data for Flow smoothing}. 
For training the flow smoothing network, we need to generate unstable videos with small FOV. Specifically, for the stable video, we randomly generate a series of cropping mask. The cropped stable video will be jittered by random homography transformations. Then, we obtain a cropped unstable video for training and the cropped stable video for supervision.

\noindent\textbf{Data for Flow Outpainting}. 
To supervise the learning of large-FOV optical flow fields, we mask the boundaries of stable videos. Specifically, we set up a sliding window $640 \times 360$, which moves randomly with the video timeline. Then, we obtain a cropped video for training and the corresponding full-frame video for supervision.

\section{Implementation Details} \label{subsec:details}

We will illustrate the training details of different networks, including the camera pose regression network, the optical flow smoothing network, and the flow outpainting network. All networks are implemented using Pytorch.

\noindent{\textbf{Camera pose regression network}}. We first describe the architecture of the camera pose regression network. The network processes each input concatenated tensor $f_{in} \in \mathbb{R}^{b \times 3\times h\times w}$ with several $2$D convolutional layers, where $b$ indicates the batch dimension and $h \times w$ indicates the spatial dimensions. The final predicted parameters are obtained by a series of $1$D convolutional layers. 
We use a batch size of $40$ and train for $10k$ iterations. we use Adam optimizer~\cite{adam} with a constant leaning rate of $10^{-4}$ for the first $4k$ iterations, followed by an exponential decay of $0.99995$ until iteration $10k$. The input resolution is set to $256 \times 512$. The weights in training loss Eq.~(5) and Eq.~(7) in the main paper are set to $\lambda_{\theta}=1.0, \lambda_{s}=1.0, \lambda_{t}=1.5, \lambda_{grid}=2.0$ for the first $6k$ iterations and $\lambda_{\theta}=2.0, \lambda_{s}=8.0, \lambda_{t}=1.0, \lambda_{grid}=2.0$ for the remaining $4k$ iterations.

\vspace{10pt}
\noindent{\textbf{Optical flow smoothing network}}. We use a batch size of $6$ and train for $20k$ iterations. we use Adam optimizer~\cite{adam} with a constant leaning rate of $10^{-4}$ for the first $10k$ iterations, followed by an exponential decay of $0.99995$ until iteration $20k$. The input resolution is set to $488 \times 768$.

\vspace{10pt}
\noindent{\textbf{Flow outpainting network}}. We apply an Unet architecture with gated convolution layers~\cite{gated_conv} as a flow-outpainting network. We use a batch size of $12$ and train for $20k$ iterations. we use the Adam optimizer~\cite{adam} with a constant leaning rate of $10^{-4}$. The input resolution is set to $488 \times 768$. The weights in training loss Eq.~(14) in the main article are set to $\lambda_{in}=2.0, \lambda_{out}=1.0, \lambda_{F}=10.0$ for the first $10k$ iterations and $\lambda_{in}=0.6, \lambda_{out}=1.0, \lambda_{F}=0.0$ for the remaining $10k$ iterations.

\section{Qualitative Evaluation} \label{subsec:qualitative evaluation}

\begin{figure*}[!h]
  \centering
  \includegraphics[width=\linewidth]{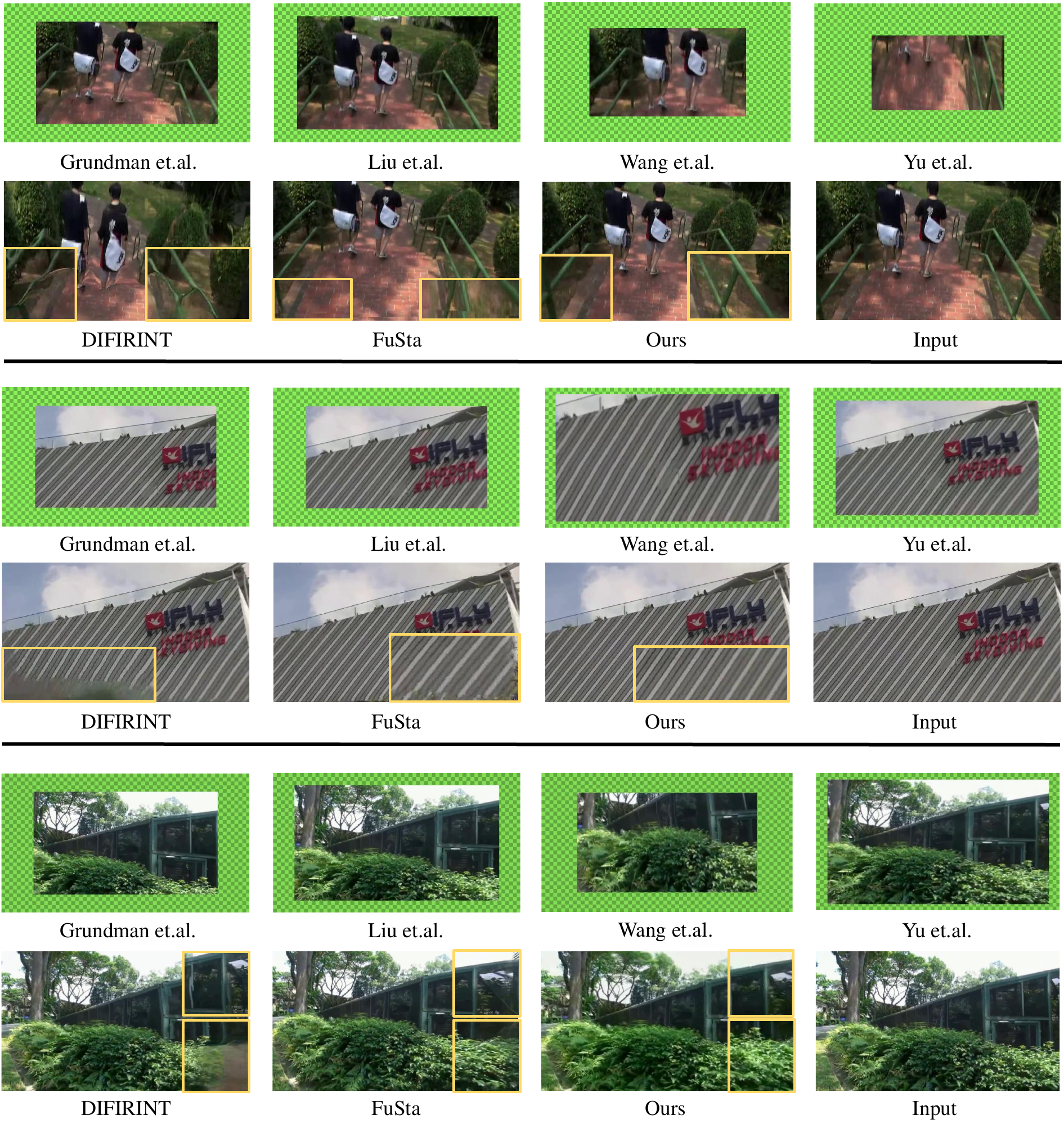}
  \caption{\textbf{Visual comparison to state-of-the-art methods.} Our proposed method does not suffer from aggressive cropping of frame borders~\cite{L1,Bundle,Wang19,yu20} and rendering artifacts than DIFRINT~\cite{DIFRINT} and FuSta~\cite{FuSta}. Specially, we keep more of the content in the input frames than FuSta~\cite{FuSta}. }
  \label{fig:vis_compare}
\end{figure*}
We show the results of the comparison of our method and the latest approaches in Fig.~\ref{fig:vis_compare}. Most methods~\cite{L1,Bundle,Wang19,yu20} suffer from a large amount of cropping, as indicated by the green checkerboard regions. Compared to full frame rendering approaches for interpolation~\cite{DIFRINT} / generation~\cite{FuSta}, our method shows fewer visual artifacts. In particular, FuSta~\cite{FuSta} would discard most of the input frame content for stabilization and deblurring, while we argue that video stabilization is based on destroying as little of the input frame content as possible. Thus, our method preserves the original content of the input frame as much as possible. We strongly recommend that the reviewers see our \textcolor{red}{\textit{additional supplementary video}}, especially \textit{the comparison with other full-frame approaches} (FuSta~\cite{FuSta}, DIFRINT~\cite{DIFRINT}).

\section{More Experimental Results}

\vspace{10pt}
\noindent \textbf{Per-category Evaluation.} We present the the average scores for the $6$ categories in the NUS dataset~\cite{Bundle}.

\begin{figure}[!h]
  \centering
  \includegraphics[width=0.90\linewidth]{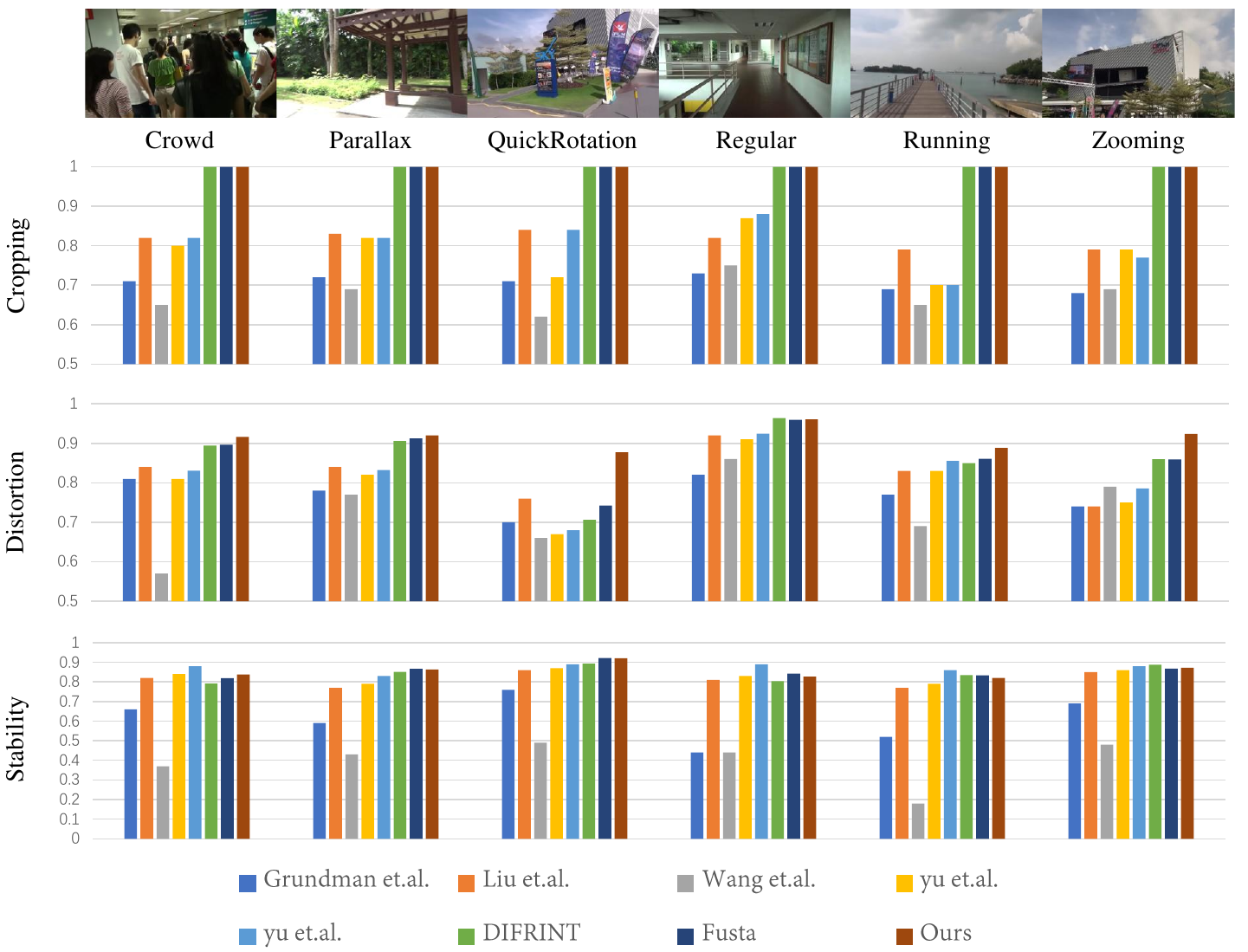}
  \caption{\textbf{Per-category quantitative evaluation on NUS dataset.} We compare the cropping ratio, distortion value, and stability score with state-of-the-art methods~\cite{L1,Bundle,Wang19,yu19,yu20,FuSta,DIFRINT}.}
  \label{fig:experiment_result}
\end{figure}

\begin{figure}[!h]
 \centering
 \includegraphics[width=0.95\linewidth]{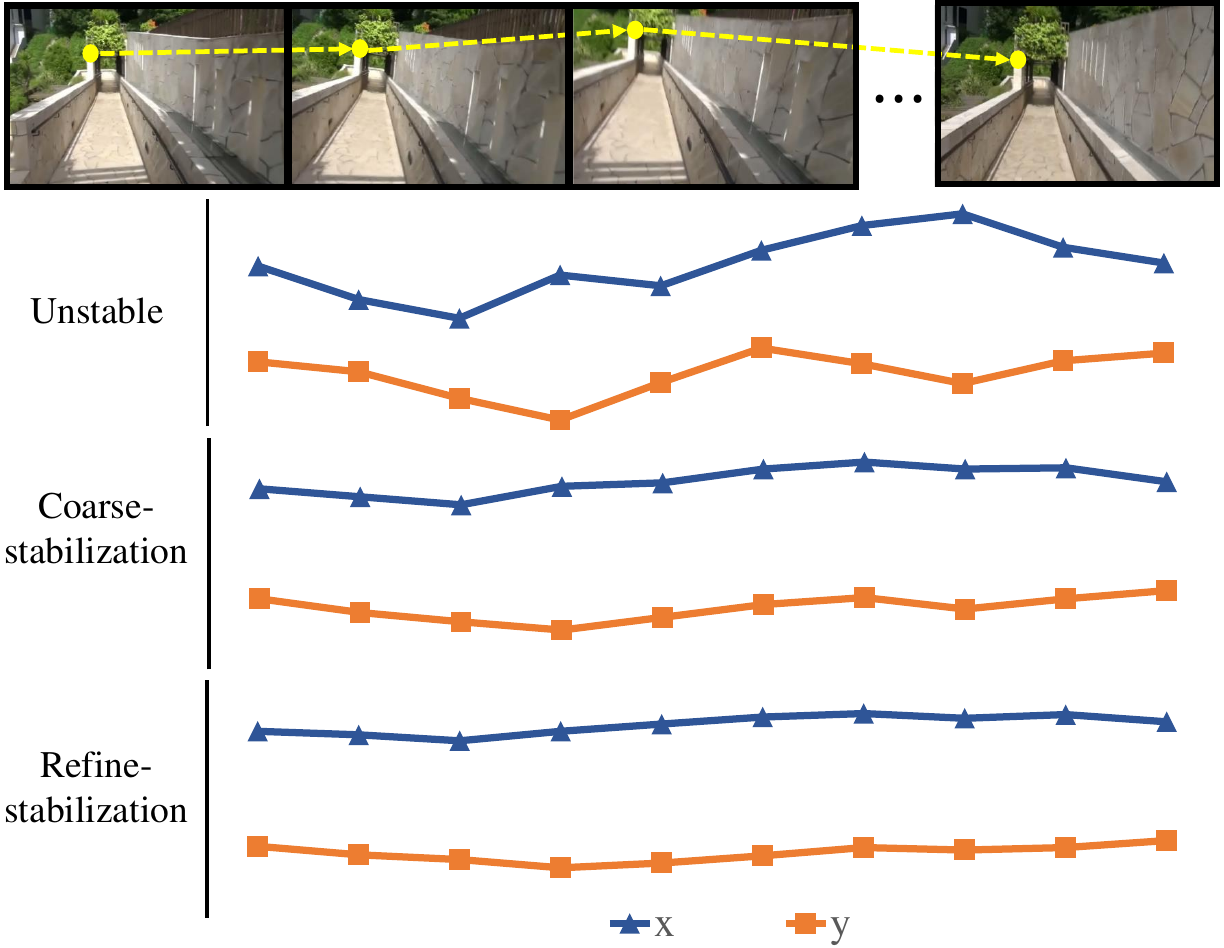}
 \caption{\textbf{Illustration of our iterative optimization-based stabilization algorithm.}}
 \label{fig:trajectory_motion}
\end{figure}

\noindent \textbf{Two-stage Stabilization.} To illustrate our two-stage stabilization method, we conduct an interesting experiment. We tracked the position $(x,y)$ of a fixed keypoint in $10$ frames, where every two frames were spaced $5$ frames apart. As shown in Fig.~\ref{fig:trajectory_motion}, the trajectory of the shaky keypoint converges to a fixed/stable position through two-stage stabilization.

\noindent \textbf{Analysis of Runtime.} We attribute the faster runtime of our approach against FuSta to the following three reasons: \romannumeral1) The traditional pose regression algorithm used in FuSta is $10$ times slower than our proposed pose regression network (see Section~6.4); \romannumeral2) Our method only requires computing optical flow once per frame, while FuSta requires computing it three times and relies on additional task-specific optimization and manual adjustments (see Section~6.4); \romannumeral3) In the rendering stage, FuSta takes input from $11$ RGB frames and their corresponding optical flow, whereas our approach only requires $7$ frames. We will highlight these reasons in the final version of the manuscript.

\section{Network Architectures}

\vspace{10pt}
\noindent{\textbf{Camera pose regression network}}.
We first describe the architecture of the camera pose regression network. 
Given a concatenated input tensor $f_{in} \in \mathbb{R}^{3\times H \times W}$, we process it with multiple down-sampled convolution layers and flatten the output feature map to $f_{out} \in \mathbb{R}^{d\times \frac{HW}{D\times D}}$\,, where $d,D$ denotes the dimension of the feature channel and the spatial down-sampling ratio, respectively. The feature vector $f_{sum}$, obtained by weighting the sum of $f_{out}$ along the feature channel, regresses all parameters of the affine transformation.
given by
\begin{equation}
    w = \psi(f_{out}), f_{sum} = \sum_{i=0}^{\frac{H W}{D\times D}}w_i f_{out}(i,\cdot), \\
    \{\theta,s,d_x,d_y\}=\mho(f_{sum})\,.
\end{equation}
Specifically, The network processes each input concatenated tensor $f_{in} \in \mathbb{R}^{b \times 3\times h\times w}$ with several $2$D convolutional layers, as shown in Table~\ref{table:arch-1}, where $b$ indicates the batch dimension and $h \times w$ indicate the spatial dimensions. The final predicted parameters are obtained by a series of $1$D convolutional layers. 

\begin{table}[!h] \footnotesize
\centering
\addtolength{\tabcolsep}{1pt}
\caption{Modular architecture of camera pose regression network modules. Each convolution operator is followed by batch normalization and LeakyReLU (negative\_slope=$0.1$), except for the last one. $K$ refers to the kernel size, $s$ denotes the stride, and $p$ indicates the padding. We apply the Max-pooling layer to downsample each feature map.}
\label{table:arch-1}
\begin{tabular}{@{}l|l|l@{}}
\toprule 
Input Size                                          &Convolution Layer                                 &Output Size\\ 
& ($K\times K$, $s$, $p$) & \\ \midrule
Feature map extraction \\ \midrule
input: $b \times 3\times h\times w$  & conv0: ($3\times 3, 1, 1$)           &$b \times 8\times h\times w$\\ \midrule
conv0: $b \times 8\times h\times w$  & conv1: ($3\times 3, 1, 1$)         &$b \times 32\times h\times w$\\ \midrule
conv1: $b \times 32\times h\times w$  & pool1: ($5\times 5, 2, 4$)      &$b \times 32\times \frac{h}{4} \times \frac{w}{4}$\\ \midrule
pool1: $b \times 32\times \frac{h}{4} \times \frac{w}{4}$  & conv2: ($3\times 3, 1, 1$)         &$b \times 64\times \frac{h}{4} \times \frac{w}{4}$\\ \midrule
conv2: $b \times 64\times \frac{h}{4} \times \frac{w}{4}$  & pool2: ($5\times 5, 2, 4$)      &$b \times 64\times \frac{h}{16} \times \frac{w}{16}$\\ \midrule
pool2: $b \times 64\times \frac{h}{16} \times \frac{w}{16}$  & conv3: ($3\times 3, 1, 1$)         &$b \times 64\times \frac{h}{16} \times \frac{w}{16}$\\ \midrule

Camera pose regression  \\ \midrule
input: $b \times 64\times 1 $  & conv1: ($1, 1, 0$)      &$b \times 32\times 1$\\ \midrule
conv1: $b \times 32\times 1 $  & conv2: ($1, 1, 0$)      &$b \times 16 \times 1$\\ \midrule
conv2: $b \times 16\times 1 $  & conv3: ($1, 1, 0$)      &$b \times 4 \times 1$\\
\bottomrule
\end{tabular}
\end{table}

\vspace{10pt}
\noindent{\textbf{Flow outpainting network}}. We apply a Unet architecture with gated convolution layers~\cite{gated_conv} as a flow outpainting network, as shown in Table~\ref{table:arch-2}.

\begin{table}[!h] \scriptsize
\centering
\addtolength{\tabcolsep}{1pt}
\caption{Architecture of the flow-outpainting network. Each $2$D gated-convolution~\cite{gated_conv} (`G\_conv') is followed by batch normalization and Sigmoid. The final `conv' denotes the $2$D convolution layer without batch normalization and Sigmoid. $K$ refers to the kernel size, $s$ denotes the stride, and $p$ indicates the padding. We apply the Maxpooling Layer for downsampling (`down') and bilinear interpolation for upsampling (`up').}
\label{table:arch-2}
\begin{tabular}{@{}l|l|l@{}}
\toprule 
Input Size                                          &Convolution Layer                                 &Output Size\\ 
& ($K\times K$, $s$, $p$) & \\ \midrule
input: $b \times 3\times h\times w$  & down\_0   &$b \times 3\times \frac{h}{4}\times \frac{w}{4}$\\ \midrule
down\_0: $b \times 3\times \frac{h}{4}\times \frac{w}{4}$  & G\_conv0: ($3\times 3, 1, 1$)   &$b \times 16\times \frac{h}{4}\times \frac{w}{4}$\\ \midrule
G\_conv0: $b \times 16\times \frac{h}{4}\times \frac{w}{4}$  & down\_1         &$b \times 16\times \frac{h}{8}\times \frac{w}{8}$\\ \midrule
down\_1: $b \times 16\times \frac{h}{8}\times \frac{w}{8}$  & G\_conv1: ($3\times 3, 1, 1$)   &$b \times 64\times \frac{h}{8}\times \frac{w}{8}$\\ \midrule
G\_conv1: $b \times 64\times \frac{h}{4}\times \frac{w}{4}$  & down\_2         &$b \times 64\times \frac{h}{16}\times \frac{w}{16}$\\ \midrule
down\_2: $b \times 64\times \frac{h}{16}\times \frac{w}{16}$  & G\_conv2: ($3\times 3, 1, 1$)   &$b \times 64\times \frac{h}{16}\times \frac{w}{16}$\\ \midrule
G\_conv2: $b \times 64\times \frac{h}{16}\times \frac{w}{16}$  & conv0: ($3\times 3, 1, 1$)  &$b \times 64\times \frac{h}{16}\times \frac{w}{16}$\\ \midrule
conv0: $b \times 64\times \frac{h}{16}\times \frac{w}{16}$  & G\_conv3: ($3\times 3, 1, 1$)  &$b \times 32\times \frac{h}{16}\times \frac{w}{16}$\\ \midrule
G\_conv3: $b \times 32\times \frac{h}{16}\times \frac{w}{16}$  & up\_0  &$b \times 32\times \frac{h}{8}\times \frac{w}{8}$\\ \midrule
up\_0+G\_conv1: $b \times 96\times \frac{h}{8}\times \frac{w}{8}$  & G\_conv4: ($3\times 3, 1, 1$)  &$b \times 16\times \frac{h}{8}\times \frac{w}{8}$\\ \midrule
G\_conv4: $b \times 16\times \frac{h}{8}\times \frac{w}{8}$  & up\_1  &$b \times 16\times \frac{h}{4}\times \frac{w}{4}$\\ \midrule
up\_1+G\_conv0: $b \times 32\times \frac{h}{4}\times \frac{w}{4}$  & conv0: ($3\times 3, 1, 1$)  &$b \times 2\times \frac{h}{4}\times \frac{w}{4}$\\ \midrule
conv0: $b \times 2\times \frac{h}{4}\times \frac{w}{4}$  & up\_2  &$b \times 2\times h\times w$\\
\bottomrule
\end{tabular}
\end{table}

\section{Limitations}
Although our method achieves a comparable stability score, we use only a simple Gaussian sliding window filter to smooth the camera trajectory in the coarse stage, leaving room for further improvement. In addition, our rendering strategy could generate artifacts in human-dense scenarios due to the nonrigid transformation of the human body, breaking our assumption of local spatial coherence. 

\end{appendices}

\end{document}